\documentclass[a4paper,12pt]{article}

\usepackage{comment}
\usepackage{fancyvrb}
\usepackage{graphicx}
\usepackage{hyperref}
\usepackage[ragged]{footmisc}
\usepackage{ragged2e}
\usepackage[section]{placeins}
\usepackage{geometry}
\usepackage{appendix}

\begin{document}

\title{Interpreting Winograd Schemas Via the SP Theory of Intelligence and Its Realisation in the SP Computer Model}

\author{J Gerard Wolff\footnote{Dr Gerry Wolff, BA (Cantab), PhD (Wales), CEng, MBCS, MIEEE; CognitionResearch.org, Menai Bridge, UK; \href{mailto:jgw@cognitionresearch.org}{jgw@cognitionresearch.org}; +44 (0) 1248 712962; +44 (0) 7746 290775; {\em Skype}: gerry.wolff; {\em Web}: \href{http://www.cognitionresearch.org}{www.cognitionresearch.org}.}}

\maketitle

\begin{abstract}

In `Winograd Schema' (WS) sentences like {\em The city councilmen refused the demonstrators a permit because they feared violence} and {\em The city councilmen refused the demonstrators a permit because they advocated revolution}, it is easy for adults to understand what ``they'' refers to but can be difficult for AI systems. This paper describes how the SP System---outlined in an appendix---may solve this kind of problem of interpretation. The central idea is that a knowledge of discontinuous associations amongst linguistic features, and an ability to recognise such patterns of associations, provides a robust means of determining what a pronoun like ``they'' refers to. For any AI system to solve this kind of problem, it needs appropriate knowledge of relevant syntax and semantics which, ideally, it should learn for itself. Although the SP System has some strengths in unsupervised learning, its capabilities in this area are not yet good enough to learn the kind of knowledge needed to interpret WS examples, so it must be supplied with such knowledge at the outset. However, its existing strengths in unsupervised learning suggest that it has potential to learn the kind of knowledge needed for the interpretation of WS examples. In particular, it has potential to learn the kind of discontinuous association of linguistic features mentioned earlier.

\end{abstract}

{\em Keywords:}

\section{Introduction}

This paper is about `Winograd Schemas', to be described, and how they may be interpreted by means of the `SP System', to be described.

In a `background' section, next: there is first a description of the concept of a Winograd Schema; then the SP System is introduced, with a link to Appendix \ref{outline_of_sp_system_appendix} where a fuller description may be found; lastly, there is a selective review of related research, with remarks about the significance of Winograd Schemas in AI and, more specifically, in the study of `commonsense reasoning' and `commonsense knowledge'.

The main body of the paper (Section \ref{examples_section}) presents three Winograd Schemas, each with a demonstration of how it may be interpreted by means of the SP System.

\section{Background}

\subsection{The concept of a Winograd Schema}\label{winograd_schema_section}

In a paper published in 1972 \cite{winograd_1972}, Terry Winograd gave an example of a pair of sentences:

\begin{quote}

{\em The city councilmen refused the demonstrators a permit because they feared violence} and

{\em The city councilmen refused the demonstrators a permit because they advocated revolution}

\end{quote}

\noindent which are easy for people to understand but are problematic for most AI systems---because the meaning of the word ``they'' in each sentence depends on a ``sophisticated knowledge of councilmen, demonstrators, and politics'' (p.~33).

Since then, researchers have gathered together many such pairings which, in a similar way, differ in only one or two words, that contain an ambiguity that is resolved differently in the two cases, and that requires the use of world knowledge and reasoning for its resolution. Some of those pairings---instances of the afore-mentioned ``Winograd Schema''---may be seen on \href{https://bit.ly/2MPm64B}{bit.ly/2MPm64B}.

For the sake of brevity in this paper, `Winograd Schema' is shortened to `WS' and `Winograd Schema Challenge' is shortened to `WSC'. Since each item in a WS pairing may comprise two or more sentences, it is wrong to refer to them as WS sentences. Instead, they will be called `WS items' or `WSIs'.

Important features of any WSI include:

\begin{itemize}

    \item ``It should be easily disambiguated by the human reader. Ideally, this should be so easy that the reader does not even notice that there is an ambiguity ...'' \cite[p.~557]{levesque_etal_2012}.

    \item It should not be possible to interpret a candidate WSI correctly using ``cheap tricks'' {\em aka} ``heuristics'' \cite[Section 2.2]{levesque_2014}. For example, with a question like {\em Could a crocodile run a steeplechase?}, we can say ``no'' because we have never heard of such a thing, without bothering to consider the physical demands of a steeplechase, the short legs and aquatic habits of crocodiles, and so on.

        In a similar way, with a pair of sentences like these: {\em The women stopped taking the pills because they were [pregnant/carcinogenic]}, the meaning of ``they'' is obvious in each case because only women can be pregnant, and only pills would be carcinogenic \cite[Section `Pitfall 1']{levesque_2011}.

    \item The test should be {\em Google-proof}. With respect to the previous example, Levesque says ``In linguistics terminology, the anaphoric reference can be resolved using selectional restrictions alone. Because selectional restrictions like this might be learned by sampling a large enough corpus (that is, by confirming that the word ``pregnant'' occurs much more often close to ``women'' than close to ``pills''), we should avoid this sort of question.'' ({\em ibid.}).

    \item There should not be too much ambiguity about the answers, as, for example, in: {\em Frank was pleased when Bill said that he was the winner
        of the competition}. Here, the word ``he'' could refer equally well to Frank or Bill \cite[Section `Pitfall 2']{levesque_2011}.

\end{itemize}

A comprehensive solution to the problems posed by WSIs depends on two things: 1) An ability to learn relevant syntax and semantics, including knowledge of the world like the ``sophisticated knowledge of councilmen, demonstrators, and politics'' noted by Winograd, and 2) An ability to make use of that syntactic and semantic knowledge in the interpretation of WSIs.

For the avoidance of any misunderstanding, this paper concentrates on the second aspect of the problem, although it does have something to say (in Sections \ref{the_sp_system_section} and \ref{unsupervised_learning_potential_section}, and in Appendix \ref{unsupervised_learning_appendix}), about the learning of syntax and semantics and how that learning relates to WSs and their interpretation.

\subsection{The {\em SP System}}\label{the_sp_system_section}

The {\em SP System}---meaning the {\em SP Theory of Intelligence} and its realisation in terms of the {\em SP Computer Model}---has been under development since about 1987, with a break between early 2006 and late 2012. There is an outline of the SP System in Appendix \ref{outline_of_sp_system_appendix}. Since an understanding of the evidence and arguments presented in this paper depend on an understanding of the workings of the SP System, {\em readers who are not already familiar with the system are urged to read Appendix \ref{outline_of_sp_system_appendix} before reading the rest of the paper.}

As noted in Appendix \ref{outline_of_sp_system_appendix}, the overarching goal in the development of the SP System has been to create a system that can simplify and integrate observations and concepts across artificial intelligence, mainstream computing, mathematics, and human learning, perception, and cognition.

To a large extent, this quest has been successful, as outlined in Appendix \ref{outline_of_sp_system_appendix}. The SP System combines relative simplicity with versatility in diverse aspects of intelligence (Appendix \ref{versatility_in_intelligence_appendix}), versatility in the representation of diverse kinds of knowledge (Appendix \ref{versatility_in_representation_of_knowledge_appendix}), and the seamless integration of diverse aspects of intelligence and diverse kinds of knowledge in any combination (Appendix \ref{seamless_integration_appendix}). And it has several potential benefits and applications (Appendix \ref{benefits_and_applications_appendix}).

As indicated in Section \ref{winograd_schema_section}, the main focus of this paper is on how the SP System may function in the interpretation of WSs, with relatively little to say about the learning of relevant syntax and semantics. However, for reasons given in Appendix \ref{unsupervised_learning_appendix}, it is anticipated that, when the learning capabilities of the SP Computer Model have been further developed for the unsupervised learning of natural language syntax, those capabilities will generalise with little or no modification to the learning of semantic structures and the learning of structures in which syntax and semantics are integrated.

\subsubsection{Evaluation of the SP System in terms of {\em simplicity} and {\em power}}

A major strength of the SP System is the way that it combines conceptual simplicity with high levels of explanatory and descriptive power (Appendix \ref{strengths_potential_of_sp_system_appendix}). This has a bearing on how the system may be evaluated in terms of its performance in the interpretation of WSs.

As we shall see with the examples in Sections \ref{city_councilmen_section}, \ref{pete_envies_martin_section}, and \ref{fish_ate_worm_section}, the SP System, with appropriate data, can provide robust interpretations of WSs, in accordance with what people judge to be correct. These interpretations are achieved with the SP Computer Model as it was developed to meet its original objectives, and without any adaptation or modification of any kind.

This means that, in terms of `simplicity' and `power' (Appendix \ref{ic_simplicity_power_appendix}), the discovery that the SP System can achieve robust interpretations of WSs without any adaptation or modification means that the simplicity of the system is the same as before, but (our knowledge of) its descriptive and explanatory power has been increased. Thus, in terms of Ockham's razor, and in terms of our knowledge of the system and what it can do, the SP System is more plausible than before.

By contrast, any system that has been developed as an {\em ad hoc} solution to the problem of interpreting WSs is likely to be much less plausible than the SP system because it will not compete with the SP system in other aspects of AI.

\subsection{Related research}

There is now a large amount written about WSs, too much to be described exhaustively in this paper. Hence, this section presents a selective review, concentrating on more recent studies.

Hector Levesque \cite{levesque_2011} presents an alternative to the Turing Test that involves responding to typed English sentences, like the Turing Test. But, unlike the Turing Test, the AI taking the test is not required to engage in a conversation and fool an interrogator into believing that he or she is dealing with a person. Instead, this WSC requires the AI to disambiguate WSIs. Levesque suggests that it is reasonable to believe that: ``with a very high probability, anything that answers [a WSC] correctly is engaging in behaviour that we would say shows thinking in people.'' and he adds: ``Whether or not a subject that passes the test is really and truly thinking is the philosophical question that Turing sidesteps.'' ({\em ibid.}, Discussion and Conclusion).

There is more about the WSC and associated issues in \cite{levesque_etal_2012}. In that paper, the authors make a connection between the study of WSs and the study of commonsense knowledge (and, by implication, commonsense reasoning): ``...~we believe that in order to pass the [WSC], a system will need to have commonsense knowledge about space, time, physical reasoning, emotions, social constructs, and a wide variety of other domains. Indeed, we hope that the [WSC] will spur new research into representations of commonsense knowledge.'' ({\em ibid}., p.~558). But they add: ``However, nothing in the [WSC] insists on this approach, and we would expect researchers [in natural language processing] to try different approaches.'' ({\em ibid}.).

Leora Morgenstern and Charles Ortiz \cite{morgenstern_ortiz_2015} describe the WSC competition, now run roughly once a year and sponsored by Nuance Communications, Inc., to encourage efforts to develop programs that can correctly interpret WSs. The first of these events was run on July 11, 2016, at the International Joint Conference on Artificial Intelligence (IJCAI-16). ``The first round of the challenge was a collection of 60 PDPs [Pronoun Disambiguation Problems]. The highest score achieved was 58\% correct, by Quan Liu, from University of Science and Technology, China. Hence, by the rules of that challenge, no prizes were awarded, and the challenge did not proceed to the second round.'' (from \href{https://bit.ly/2R4gSFJ}{bit.ly/2R4gSFJ}).

Altaf Rahman and Vincent Ng \cite{rahman_Ng_2012} examine the task of determining (`resolving') the meanings of pronouns in sentences where there are no obvious clues from the syntax---much as in the WSC. They present results from a ``Combined Resolver''---a combination of the ``Stanford resolver'' and the ``Baseline Ranker''---which ``significantly outperforms state-of-the-art resolvers'' but has ``a lot of room for improvement'' ({\em ibid}., Section 5.4).

In a paper entitled ``On our best behaviour'', Hector Levesque \cite{levesque_2014} discusses some some general issues relating to the WSC including: ``...~what does it tell us when a good semblance of a behaviour can be achieved using cheap tricks that seem to have little to do with what we intuitively imagine intelligence to be?'' and ``...~are the philosophers right, and is a behavioural understanding of intelligence simply too weak?'' He suggests that both those ideas are wrong and that ``...~we should put aside any idea of tricks and short cuts, and focus instead on what needs to be {\em known}, how to represent it symbolically, and how to use the representations.'' ({\em ibid}., p.~34, emphasis in the original). Two main hurdles need to be overcome: much of what we know comes via language, but we need knowledge to interpret language; and even at the level of what is known by children, there appears to be much complexity and heavy computational demands.

With respect to the WSC, Peter Sch{\"u}ller \cite{schuller_2014} considers pairs of sentences like these: {\em Sam's drawing was hung just above Tina's  and it did look much better with another one [below / above] it}. Here, people naturally assume that ``another one'' is Sam's drawing or Tina's, although it could be a third drawing by someone else. He suggests that what people naturally assume may be explained in terms of ``Relevance Theory'' and develops these ideas with a simplified version of Roger Schank's graph framework for natural language understanding.

Daniel Bailey and colleagues \cite{bailey_etal_2015}, in discussing the problem of interpreting WSs, say that ``we treat coreference
resolution as a by-product of a general process of establishing discourse coherence: a resolution for a pronoun is acceptable if it makes the discourse `coherent'.'' (p.~17). Later, they explain that, in a sentence like {\em Joan made sure to thank Susan for all the help she had given}, the phrases {\em Joan made sure to thank Susan} and {\em Susan had given help} ``are correlated in the sense that either one would cause the hearer to view the other as more plausible.'' (p.~18). They go on to present a ``correlation calculus'' for representing and reasoning about such relationships which, although it is not founded on statistical concepts of correlation, ``Nevertheless, our correlation calculus turned out to be closely related to correlation in the sense of probability theory ...'' (p.~18).

Ali Emami and colleagues \cite{emami_etal_2018} propose a three-stage knowledge hunting method for interpreting any given WSI: 1) Perform a partial parse of the WSI to isolate the main elements of the meaning; 2) Generate queries to send to a search engine in order to extract text snippets that resemble the given WSI; 3) Obtain a set of text snippets that resemble the given WSI, and use this information to resolve the ambiguity in the WSI. They say that their method compares well with AI alternatives but is not as good as a person.

Adam Richard-Bollans and colleagues \cite{richard-bollans_etal_2018} consider the WSC as a means for highlighting problems in the field of commonsense reasoning. Approaches that they discuss include machine learning approaches and `commonsense rules'. Key challenges include: pragmatics (extra-linguistic factors, such as context, and how they allow the understanding of a speaker's intended meaning), assumptions about the world (e.g., if we throw something down to someone, that person must be below you), and the level of detail or vagueness that may be needed in formalising commonsense knowledge. They propose an alternative approach, with knowledge bases for the meanings of commonsense terms and the use of heuristics based on pragmatics.

Arpit Sharma and colleagues \cite{sharma_etal_2015a,sharma_etal_2015b} present an approach that identifies the knowledge needed to answer a challenge question, hunts down that knowledge from text repositories, and then reasons with the knowledge and the question to come up with an answer. This research includes the development of a semantic parser. They show that their approach works well on a subset of Winograd schemas.

\section{Examples of how the SP System may disambiguate WSIs}\label{examples_section}

The three subsections that follow present three pairs of WSIs, each one with an SP-multiple-alignment produced by the SP Computer Model that demonstrates how the ambiguity in the sentence may be resolved.

A feature of the SP Computer Model is that it has a tendency to produce SP-multiple-alignments that are very large and cannot easily be fitted on one page. Quite often it is possible to work around this problem by choosing suitable examples or splitting SP-multiple-alignments into two or more parts. But with these examples, even with the choice of relatively short WSIs, those work-arounds are not possible---so it has been necessary to do something different.

The solution that has been adopted with these examples is that each SP-multiple-alignment has been produced at full size in vector-graphic form in a PDF file, and then shrunk to a size that will fit on a page. Then, providing each SP-multiple-alignment is viewed on a computer, it may be magnified to whatever size makes it legible, and it will be sharp at any size because of the vector-graphic encoding. For the convenience of readers, the figures are available in a separate file (`sp\_ws\_figures.pdf'), in addition to their display in the body of this paper. That means that the figures in the separate file may be magnified without magnifying the main body of the paper.

A point to note about these examples is that, although the words `syntax' and `semantics' are used in discussing the examples, the SP System makes no formal distinction between the two. In the SP System, both `syntax' and `semantics' are information and may be combined flexibly without reference to any informal classification we may apply to them.

It should be stressed again that, up to now in this research, there has been no attempt to apply any kind of learning processes. In each of the examples to be described, the New SP-pattern, and all the Old SP-patterns, have been supplied ready-made to the SP Computer Model (see also Section \ref{unsupervised_learning_potential_section}).

It is likely that there is much room for debate about the details of these examples. But any such concerns should not distract from the main point of the examples: that within the framework of the SP system, and with WSIs containing a pronoun that requires disambiguation, it is relatively straightforward to discover the referent of such a pronoun.

All the WSIs in what follows contain at least one capital letter. But in the sentences provided as New patterns for the SP Computer Model, all capital letters have been replaced with their lower-case equivalents. This is to by-pass rules for capitalisation which might otherwise be a distraction from the main point of these examples: how to find the referent for the pronoun in each example.

\subsection{The city councilmen refused the demonstrators a permit}\label{city_councilmen_section}

This subsection is concerned with Terry Winograd's \cite[p.~33]{winograd_1972} two example sentences:\footnote{Apart from one difference, this is the first example in the `Collection of Winograd Schemas', compiled by Ernest Davis and colleagues, and shown on \href{https://bit.ly/2MPm64B}{bit.ly/2MPm64B}, retrieved 2018-09-26. The difference is to use exactly the same two sentences presented by Winograd \cite[p.~33]{winograd_1972} which, at their ends, have [`feared violence'/`advocated revolution'] instead of [`feared violence'/`advocated violence'].}

\begin{quote}

{\em The city councilmen refused the demonstrators a permit because they feared violence} and

{\em The city councilmen refused the demonstrators a permit because they advocated revolution}

\end{quote}

In each case, the sentence is supplied as a New SP-pattern to the SP Computer Model together with a collection of Old SP-patterns which describe the syntax and semantics of the given sentence and other sentences of that kind.

\subsubsection{The city councilmen ...~feared violence}\label{feared_violence_section}

\begin{figure}[!htbp]
\centering
\includegraphics[width=0.9\textwidth]{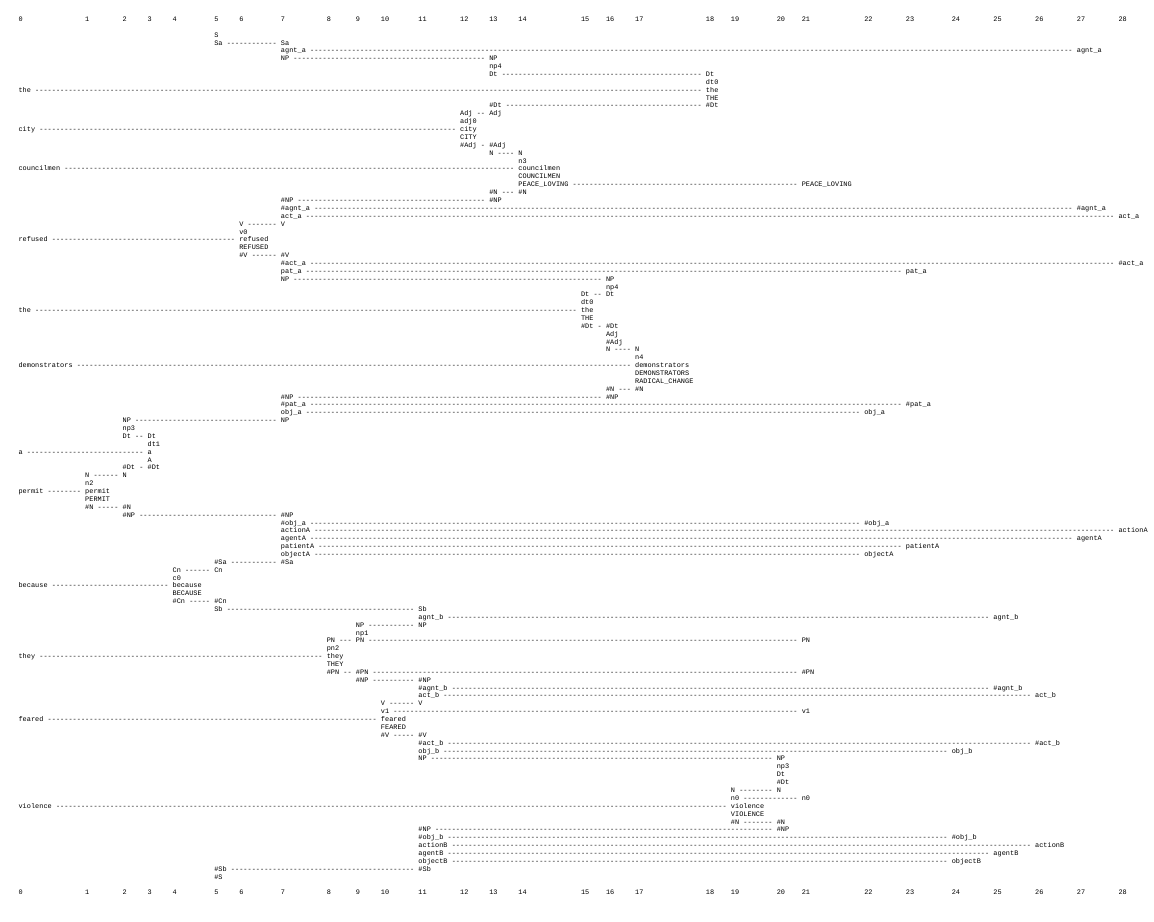}
\caption{The best SP-multiple-alignment produced by the SP Computer Model with {\em the city councilmen refused the demonstrators a permit because they feared violence} as a New SP-pattern, which appears in column 0, and a collection of Old SP-patterns as described in the text, some of which are shown in columns 1 to 28, one SP-pattern per column.}
\label{ws_councilmen_fv_figure}
\end{figure}

With the first sentence, the best SP-multiple-alignment created by the SP Computer Model is shown in Figure \ref{ws_councilmen_fv_figure}. Here, the word `best' means that, out of the many SP-multiple-alignments created by the SP Computer Model, the given SP-multiple-alignment is the one in which the New SP-pattern may be encoded most economically in terms of the Old SP-patterns in that SP-multiple-alignment, as described in \cite[Section 4.1]{sp_extended_overview} and \cite[Section 3.5]{wolff_2006}.

The SP-multiple-alignment shown in Figure \ref{ws_councilmen_fv_figure} is, if effect, a parsing of the sentence {\em the city councilmen refused the demonstrators a permit because they feared violence}, much like the parsing in Figure \ref{fortune_brave_multiple_alignment_figure}, except that SP-symbols in the sentence are whole words instead of letters as in the sentence in Figure \ref{fortune_brave_multiple_alignment_figure} and, as described in Appendix \ref{syntax_and_semantics_appendix}, the SP-multiple-alignment in Figure \ref{ws_councilmen_fv_figure}, like other examples in this paper, attempts to marry syntax with semantics. It has been moved into Appendix \ref{syntax_and_semantics_appendix} because it is quite tentative and is not relevant to the main substance of this paper---how to disambiguate pronouns in WSIs. Readers may safely ignore it if they wish.

\subsubsection{Finding what `they' refers to}\label{disambiguation_of_they_section}

Of course, the main point of interest with this example is how it determines the referent for the word ``they'' in {\em the city councilmen refused the demonstrators a permit because they feared violence}. This disambiguation is achieved via the SP-pattern `texttt{PEACE\_LOVING PN \#PN v1 n0}' which appears in column 21 in Figure \ref{ws_councilmen_fv_figure}. The connection is made correctly because, within that SP-pattern:

\begin{itemize}

    \item The SP-symbol `texttt{PEACE\_LOVING}' is aligned with the same symbol in the SP-pattern `texttt{N n3 councilmen COUNCILMEN PEACE\_LOVING \#N}' in column 14, which, in the light of the next bullet point, identifies `texttt{councilmen}' and its associated meaning as the referent for ``they'' in the given sentence.

    \item The SP-symbols `texttt{PN \#PN}' are aligned with the matching SP-symbols in `texttt{NP np1 PN \#PN \#NP}' (column 9) and in `texttt{PN pn2 they THEY \#PN}' (column 8), thus identifying the pronoun ``they''.

    \item The SP-Symbol `texttt{v1}' matches the same symbol in `texttt{V v1 feared FEARED \#V}' in column 10, thus identifying the word ``feared'' and its associated meaning.

    \item And the SP-Symbol `texttt{n0}' matches the same symbol in `texttt{N n0 violence VIOLENCE \#N}' in column 19, thus identifying the word ``violence'' and its associated meaning.

\end{itemize}

In short, within the SP-multiple-alignment shown in Figure \ref{ws_councilmen_fv_figure}, the SP-pattern `texttt{PEACE\_LOVING PN \#PN v1 n0}' has the effect of highlighting the association between peace-loving councilmen, the pronoun `they', and the councilmen's probable fear of violence.

Possible variations on this scheme are described in Section \ref{variations_section}, and the potential role for unsupervised learning is described in Section \ref{unsupervised_learning_potential_section}.

\subsubsection{Variations}\label{variations_section}

\begin{figure}[!htbp]
\centering
\includegraphics[width=0.9\textwidth]{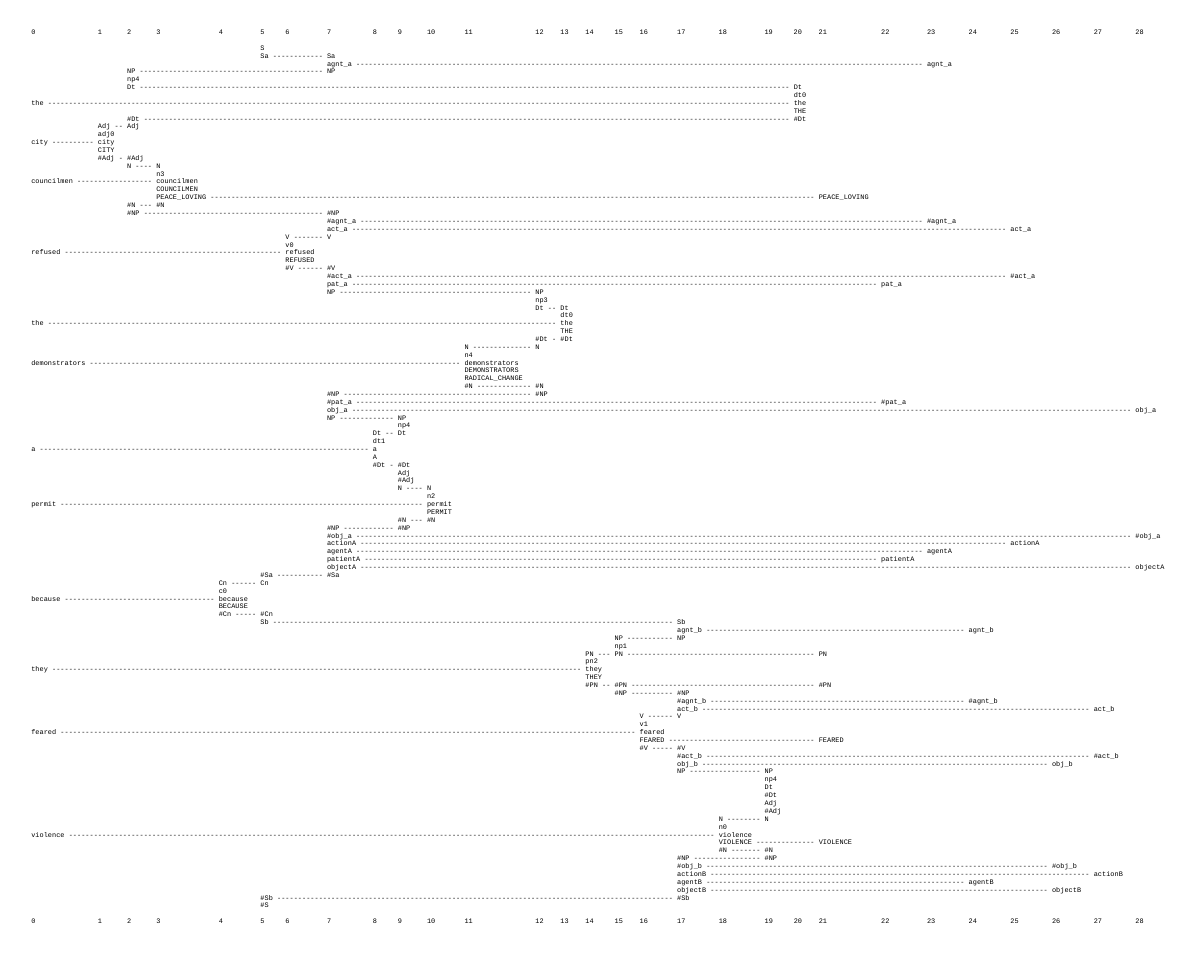}
\caption{The best SP-multiple-alignment created by the SP Computer Model with Old SP-patterns producing a variant of Figure \ref{ws_councilmen_fv_figure} as described in the text.}
\label{ws_councilmen_fv_1_figure}
\end{figure}

As mentioned near the beginning of Section \ref{examples_section}, there is plenty of room for debate about the grammatical details of these examples. Much the same goes for the proposed solution to the WS problem where several variations are possible. But, as before, any such variations should not distract from the key point: that, within the framework of the SP System, and with appropriate SP-patterns, finding the referent for a pronoun in a WSI is relatively straightforward.

As an illustration of one possible variant of the solution described in Section \ref{disambiguation_of_they_section}, Figure \ref{ws_councilmen_fv_1_figure} shows how the {\em ...~feared violence} sentence may be disambiguated via the creation of an SP-multiple-alignment, much as in Figure \ref{ws_councilmen_fv_figure}, but with the SP-pattern `texttt{PEACE\_LOVING PN \#PN FEARED VIOLENCE}' in column 21 instead of `texttt{PEACE\_LOVING PN \#PN v1 n0}'.

\begin{figure}[!htbp]
\centering
\includegraphics[width=0.9\textwidth]{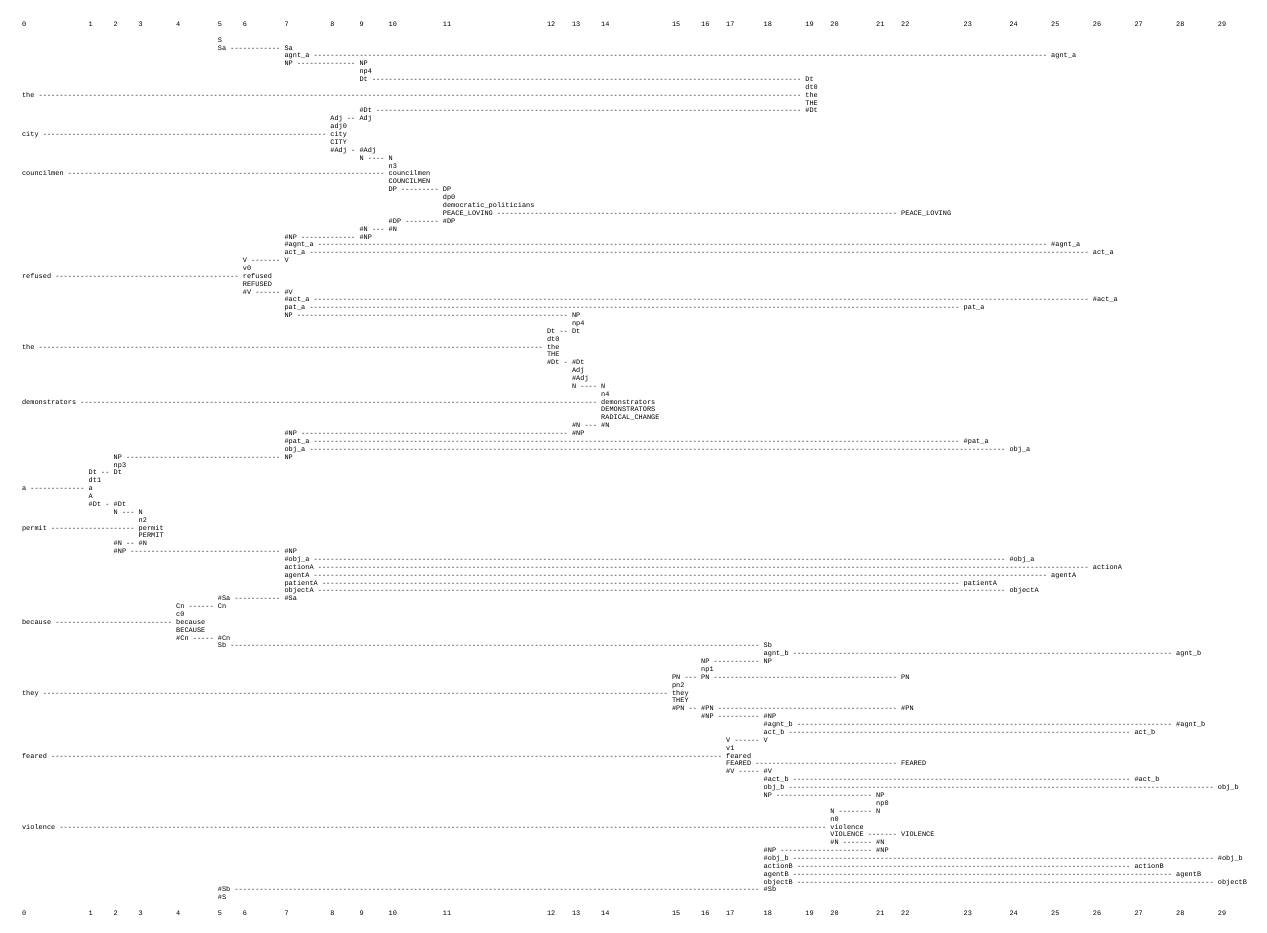}
\caption{The best SP-multiple-alignment created by the SP Computer Model with Old SP-patterns producing a variant of Figure \ref{ws_councilmen_fv_1_figure} as described in the text.}
\label{ws_councilmen_fv_3_figure}
\end{figure}

Another possible variant is shown in Figure \ref{ws_councilmen_fv_3_figure}. Here, Old SP-patterns have been supplied to the SP Computer Model which, in a simple way, introduce the concept of class-inclusion hierarchy and the inheritance of attributes.

In the figure, the `\texttt{PEACE\_LOVING}' attribute, contrary to fact, appears in a new SP-pattern for `democratic politicians', `\texttt{DP dp0 democratic\_politicians PEACE\_LOVING \#DP}', in column 11. And `\texttt{PEACE\_LOVING}' is aligned with the same symbol in the SP-pattern `\texttt{PEACE\_LOVING PN \#PN FEARED VIOLENCE}' in column 22, much as in Figure \ref{ws_councilmen_fv_1_figure}. The new SP-pattern, for `democratic politicians' connects with a modified SP-pattern for `councilmen', `\texttt{N n3 councilmen COUNCILMEN DP \#DP \#N}', in column 10. The overall effect is that, in the manner of object-oriented programming, the class `councilmen' inherits the `peace loving' attribute from the more general class `democratic politicians'.

\subsubsection{The demonstrators ...~advocated revolution}

\begin{figure}[!htbp]
\centering
\includegraphics[width=0.9\textwidth]{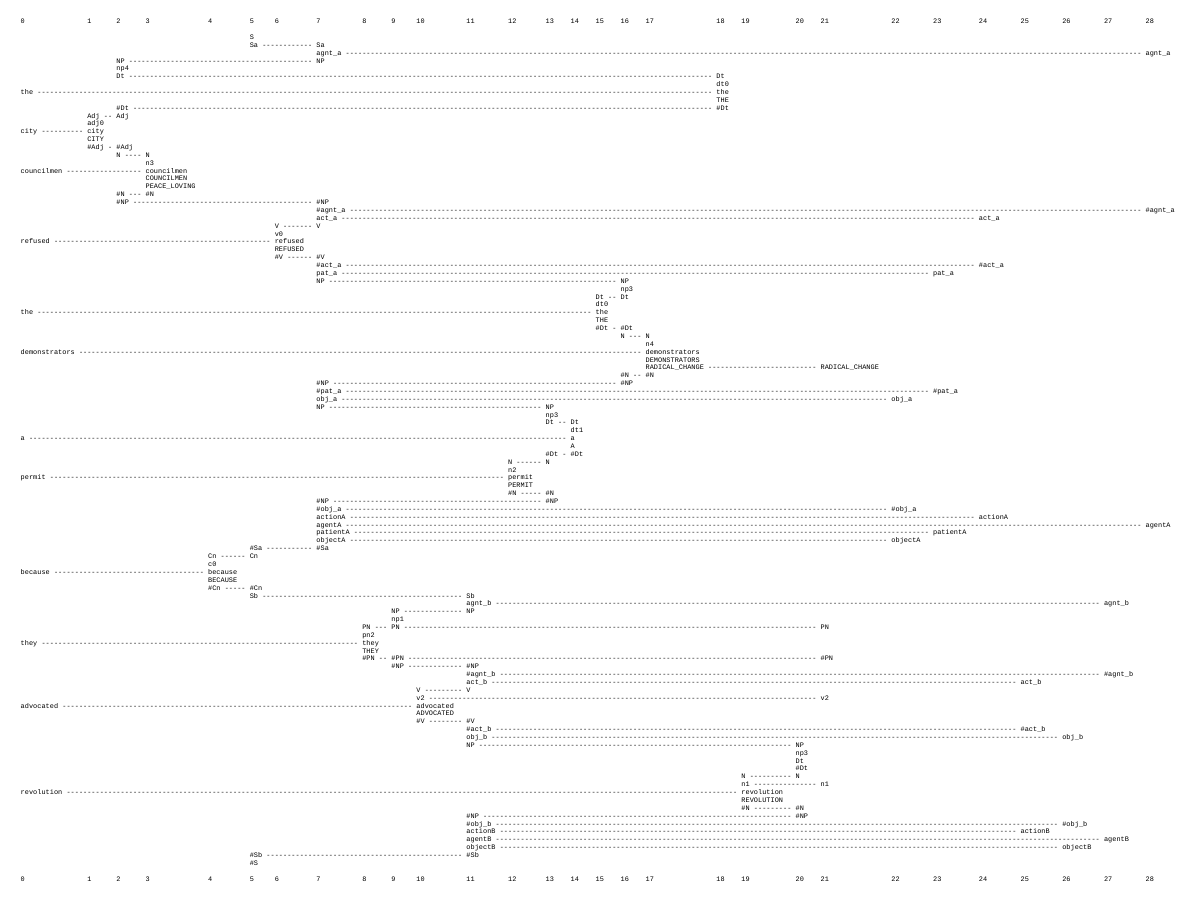}
\caption{The best SP-multiple-alignment produced by the SP Computer Model with {\em the city councilmen refused the demonstrators a permit because they advocated revolution} as a New SP-pattern, which appears in column 0, and a collection of Old SP-patterns as described in the text, some of which are shown in columns 1 to 28, one SP-pattern per column.}
\label{ws_councilmen_ar_figure}
\end{figure}

The second of Winograd's \cite{winograd_1972} WSIs, {\em The city councilmen refused the demonstrators a permit because they advocated revolution}, may be processed in a similar way. Figure \ref{ws_councilmen_ar_figure} shows the best SP-multiple-alignment created by the SP Computer Model with the given sentence as the New pattern and the same set of Old SP-patterns as before. This time, the SP-pattern `texttt{RADICAL\_CHANGE PN \#PN v2 n1}', which appears in column 21, has the effect, in much the same manner as before, of highlighting the association between the demonstrators and their probable interest in radical change, the pronoun `they', and the demonstrator's advocacy of revolution.

\subsection{Pete envies Martin}\label{pete_envies_martin_section}

The second pair of sentences to be considered are: {\em Pete envies Martin [because/although] he is very successful}.\footnote{this is the 20th example in the `Collection of Winograd Schemas', compiled by Ernest Davis and colleagues, and shown on \href{https://bit.ly/2MPm64B}{bit.ly/2MPm64B}, retrieved 2018-09-26.}

\subsubsection{Pete envies Martin because ...}\label{pete_martin_because_section}

\begin{figure}[!htbp]
\centering
\includegraphics[width=0.9\textwidth]{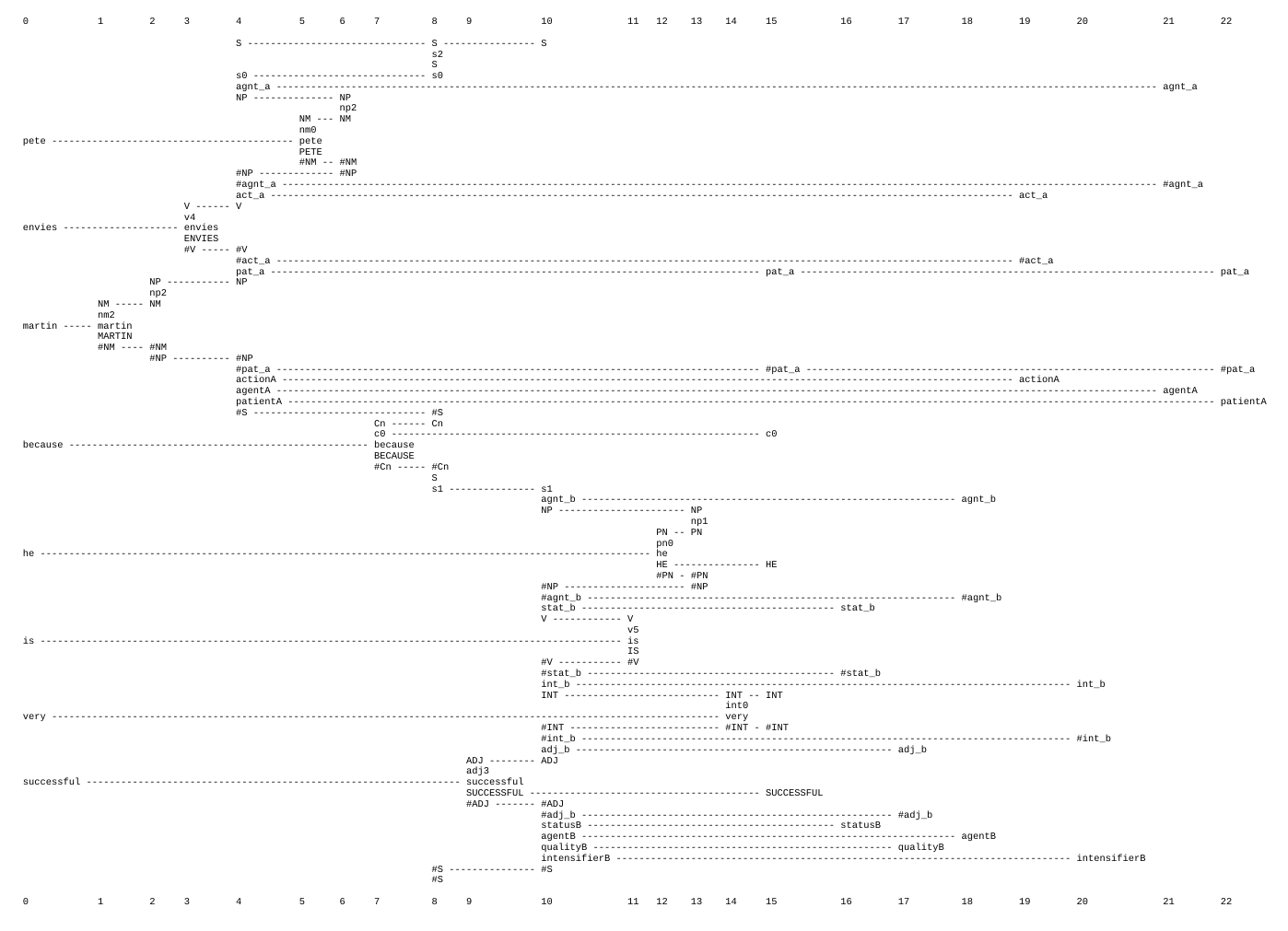}
\caption{The best SP-multiple-alignment produced by the SP Computer Model with {\em pete envies martin because he is very successful} as a New SP-pattern, which appears in column 0, and a collection of Old SP-patterns as described in the text, some of which are shown in columns 1 to 22, one SP-pattern per column.}
\label{ws_ba_b_figure}
\end{figure}

Figure \ref{ws_ba_b_figure} shows the best SP-multiple-alignment produced by the SP Computer Model with {\em pete envies martin because he is very successful} as a New SP-pattern, which appears in column 0, and a collection of Old SP-patterns describing the syntax and semantics of that kind of sentence, some of which appear in columns 1 to 22, one SP-pattern per column.

In a manner much like the examples in Section \ref{city_councilmen_section}, disambiguation is achieved via an SP-pattern that connects relevant parts of the SP-multiple-alignment. Here, that SP-pattern is `\texttt{pat\_a \#pat\_a c0 HE INT \#INT SUCCESSFUL}' which appears in column 15 in the figure. In that SP-pattern:

\begin{itemize}

    \item The pair of SP-symbols `\texttt{pat\_a \#pat\_a}', which are mnemonic for `patient', are aligned with matching symbols in the SP-pattern in column 4 which span the structure in columns 2, 1, and 0, which identify `\texttt{martin}' as having the semantic role of `patient' in the given sentence.

    \item The SP-symbol `\texttt{c0}' is aligned with the matching SP-symbol in column 7, thus identifying the special word `because' in `\texttt{Cn c0 because BECAUSE \#Cn}'.

    \item The SP-symbol `\texttt{HE}' is aligned with the matching SP-symbol within the SP-pattern `\texttt{PN pn0 he HE \#PN}' in column 12.

    \item The pair of SP-symbols `\texttt{INT \#INT}' are aligned with matching SP-symbols within the SP-pattern `\texttt{INT int0 very \#INT}' in column 14.

    \item And the SP-symbol `\texttt{SUCCESSFUL}' is aligned with the matching SP-symbol within the SP-pattern `\texttt{ADJ adj3 successful SUCCESSFUL \#ADJ}' in column 9.

\end{itemize}

In short, the SP-pattern `\texttt{pat\_a \#pat\_a c0 HE INT \#INT SUCCESSFUL}' has the effect of identifying `martin' as the referent of `he' within the given sentence, with the other SP-symbols within the SP-pattern, especially the special word `because', providing the necessary context.

\subsubsection{Pete envies Martin although ...}\label{pete_martin_although_section}

\begin{figure}[!htbp]
\centering
\includegraphics[width=0.9\textwidth]{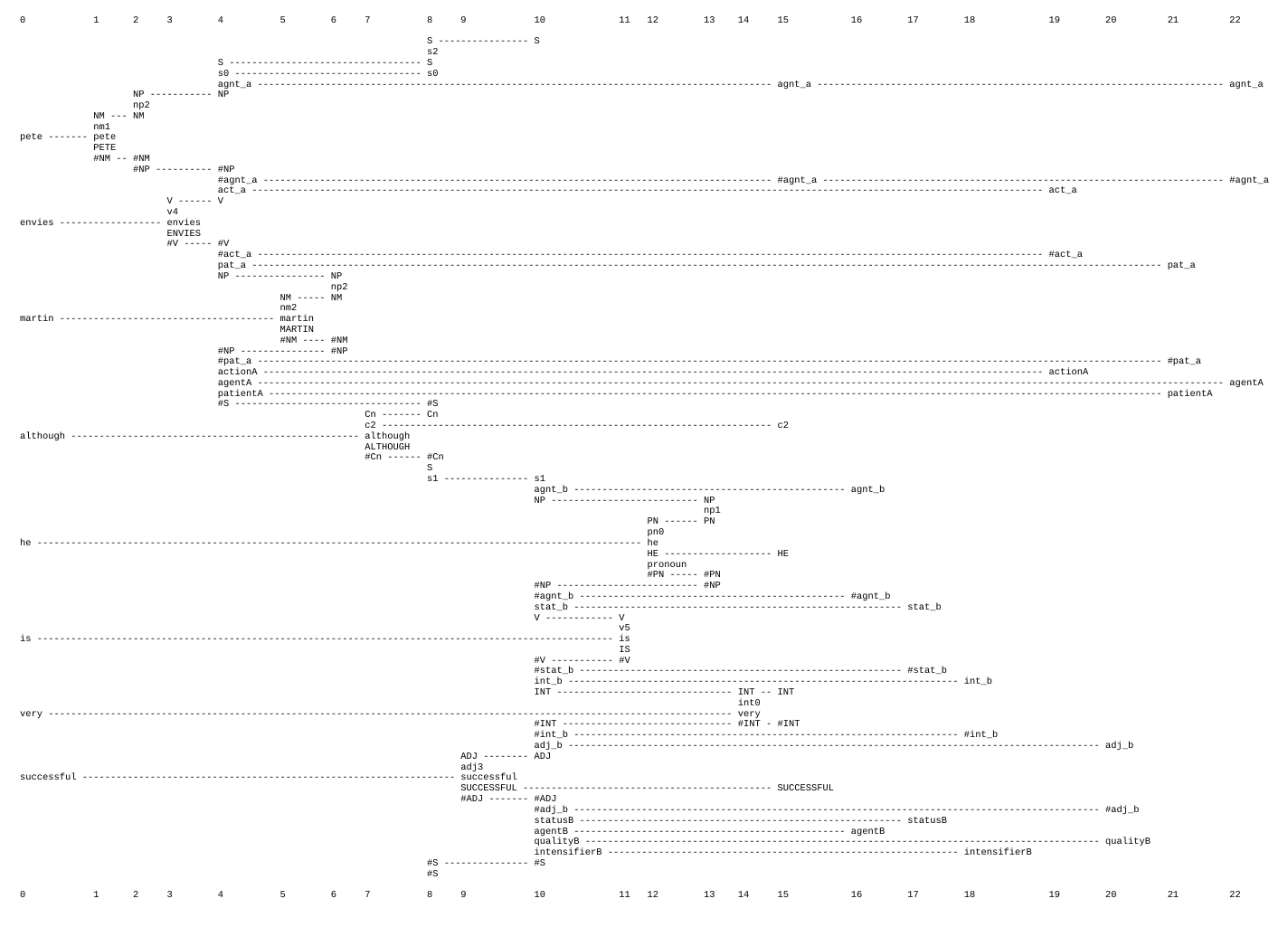}
\caption{The best SP-multiple-alignment produced by the SP Computer Model with {\em pete envies martin although he is very successful} as a New SP-pattern, which appears in column 0, and a collection of Old SP-patterns as described in the text, some of which are shown in columns 1 to 22, one SP-pattern per column.}
\label{ws_ba_a_figure}
\end{figure}

Figure \ref{ws_ba_a_figure} shows the best SP-multiple-alignment produced by the SP Computer Model with {\em pete envies martin although he is very successful} as a New SP-pattern, which appears in column 0, and a collection of Old SP-patterns describing the syntax and semantics of that kind of sentence, some of which appear in columns 1 to 22, one SP-pattern per column.

As before, there is a key SP-pattern which identifies the referent for the pronoun `he' in the given sentence. Here, that SP-pattern is `\texttt{agnt\_a \#agnt\_a c2 HE INT \#INT SUCCESSFUL}' which appears in column 15 of the figure. By contrast with the example in Section \ref{pete_martin_because_section}, the pair of SP-symbols `\texttt{agnt\_a \#agnt\_a}' are aligned with matching symbols in column 4 which have the effect of marking `\texttt{pete}' as an object of interest. In much the same way as in the example in Section \ref{pete_martin_because_section}, `\texttt{pete}' is identified as the referent of `\texttt{HE}'.

\subsection{The fish ate the worm}\label{fish_ate_worm_section}

The third example to be considered is: {\em The fish ate the worm. It was [tasty/hungry]}.\footnote{this is the 52nd example in the `Collection of Winograd Schemas', compiled by Ernest Davis and colleagues, and shown on \href{https://bit.ly/2MPm64B}{bit.ly/2MPm64B}, retrieved 2018-09-26.}

\subsubsection{The worm ...~tasty}\label{worm_tasty_section}

\begin{figure}[!htbp]
\centering
\includegraphics[width=0.9\textwidth]{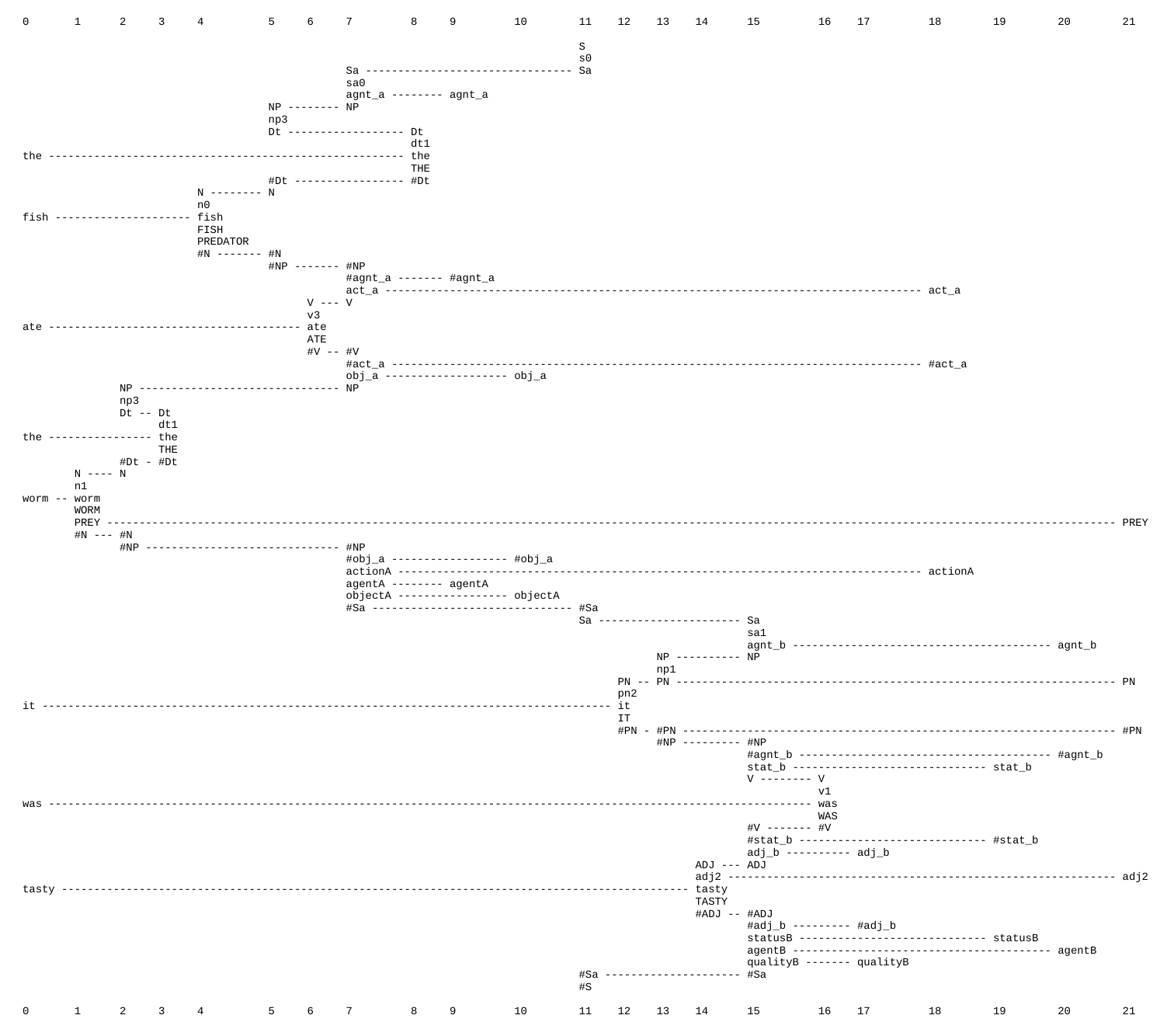}
\caption{The best SP-multiple-alignment produced by the SP Computer Model with {\em the fish ate the worm it was tasty} as a New SP-pattern, which appears in column 0, and a collection of Old SP-patterns as described in the text, some of which are shown in columns 1 to 21, one SP-pattern per column.}
\label{ws_fw_w_figure}
\end{figure}

Figure \ref{ws_fw_w_figure} shows the best SP-multiple-alignment produced by the SP Computer Model with {\em the fish ate the worm it was tasty} as a New SP-pattern, which appears in column 0, and a collection of Old SP-patterns describing the syntax and semantics of the two target sentences. Some of those SP-patterns are shown in columns 1 to 21, one SP-pattern per column.

As readers may guess, the solution in this case is similar to what has presented in previous examples. In this case, the key SP-pattern is `\texttt{PREY PN \#PN adj2}'. The first SP-symbol, `\texttt{PREY}' matches the same SP-symbol in the SP-pattern `\texttt{N n1 worm WORM PREY \#N}' in column 1, thus marking `worm' as an object of interest. The pair of SP-symbols `\texttt{PN \#PN}' match the same two SP-symbols in column 12 and in the SP-pattern `\texttt{PN pn2 it IT \#PN}' thus identifying `it' as a pronoun. And the SP-symbol `\texttt{adj2}' selects the SP-pattern `\texttt{ADJ adj2 tasty TASTY \#ADJ}.

In short, the SP-pattern `\texttt{PREY PN \#PN adj2}' tells us that the `worm' is the referent of the pronoun `it' because, as a kind of prey for animals like fish, it can be `tasty'.

\subsubsection{The fish ...~hungry}

\begin{figure}[!htbp]
\centering
\includegraphics[width=0.9\textwidth]{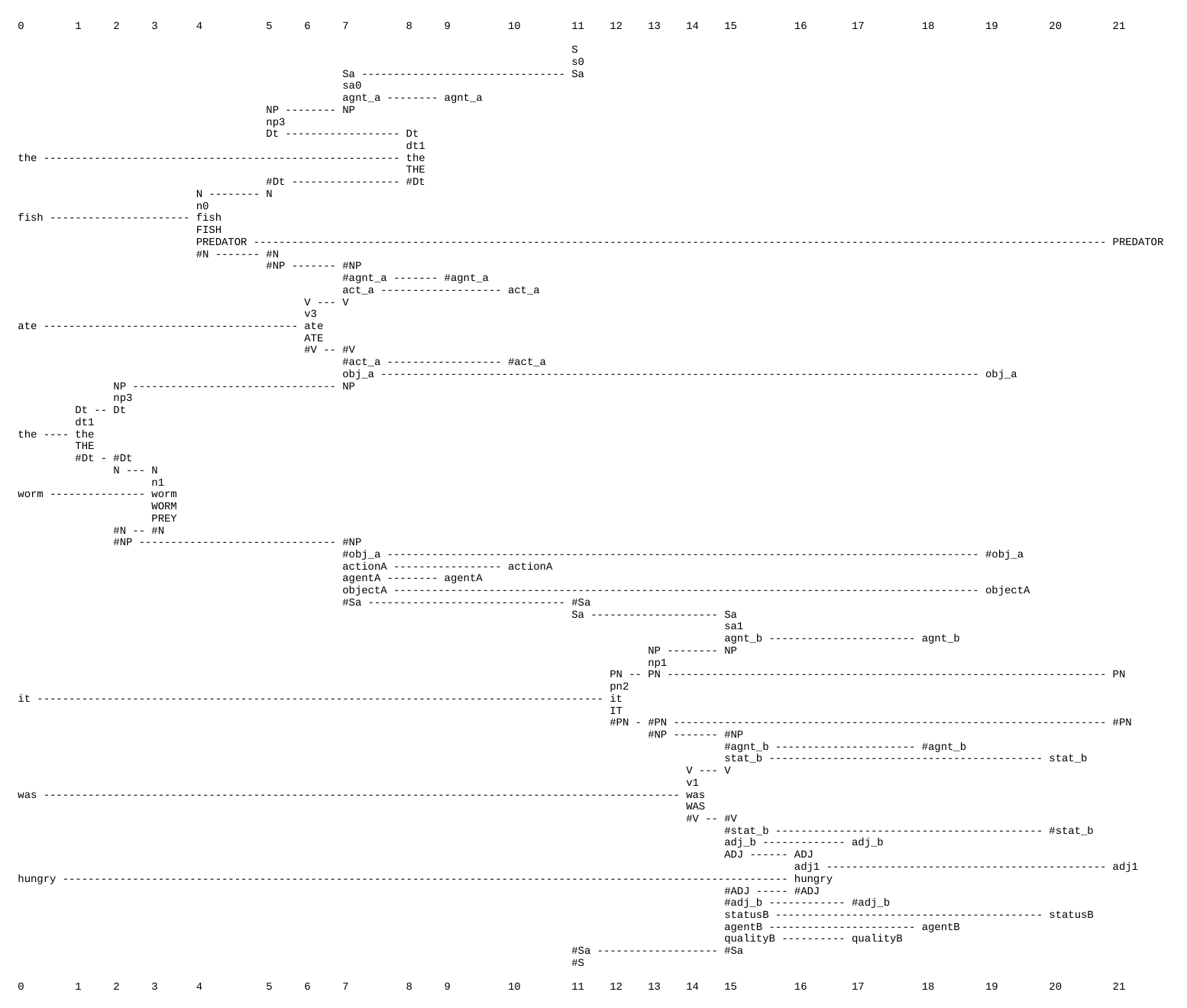}
\caption{The best SP-multiple-alignment produced by the SP Computer Model with {\em the fish ate the worm it was hungry} as a New SP-pattern, which appears in column 0, and a collection of Old SP-patterns as described in the text, some of which are shown in columns 1 to 21, one SP-pattern per column.}
\label{ws_fw_f_figure}
\end{figure}

Figure \ref{ws_fw_f_figure} shows the best SP-multiple-alignment produced by the SP Computer Model with {\em the fish ate the worm it was hungry} as a New SP-pattern, which appears in column 0, and a collection of Old SP-patterns describing the syntax and semantics of the two target sentences. Some of those SP-patterns are shown in columns 1 to 21, one SP-pattern per column.

In much the same way as in Section \ref{worm_tasty_section}, the SP-pattern `\texttt{PREDATOR PN \#PN adj1}' makes a connection between the `fish', the pronoun `it' and the fact that the fish was `hungry'.

\subsection{The potential of unsupervised learning}\label{unsupervised_learning_potential_section}

As noted in Section \ref{examples_section}, each of the SP-multiple-alignments presented in this paper has been created by the SP Computer Model with with a collection of SP-patterns supplied by the user: one New SP-pattern and a repository of Old SP-patterns including those that appear in the given SP-multiple-alignment. This is because unsupervised learning in the SP Computer Model has shortcomings (Appendix \ref{unsupervised_learning_appendix}) which mean that it is not yet good enough for demonstrating how the SP System can learn the knowledge needed for disambiguating WSIs.

However, a comprehensive account of how WSIs may be disambiguated should include an account of how relevant knowledge may be learned. In the light of what has already been learned about the strengths and weaknesses of unsupervised learning in the SP Computer Model as it is now, and how the weaknesses may be overcome (Appendix \ref{unsupervised_learning_appendix}), here are some observations about how the system may be developed to learn the kind of knowledge needed for the disambiguation of WSIs:

\begin{itemize}

    \item It is anticipated that it should be relatively straightforward to develop the SP System for the learning of segmental structures (words and phrases) and classes of such structures (e.g., nouns, verbs, and adjectives). As noted in Appendix \ref{unsupervised_learning_appendix}, the main hurdle to be overcome is the learning of structures at intermediate levels (between words and sentences) such as phrases and clauses.

    \item Likewise, it should be feasible to develop the system for the learning of discontinuous dependencies such as plurality dependencies or gender dependencies in syntax (more below). As with segmental structures, this is a weakness of the SP Computer Model as it is now (Appendix \ref{unsupervised_learning_appendix}).

    \item More challenging will be the learning of non-syntactic `semantic' knowledge of the world, mainly because it will be necessary to generalise the system for the learning of two-dimensional and three-dimensional structures, and the learning of `commonsense' knowledge of time, speed, distance, and the like.

    \item Even more challenging will be the learning of structures that combine syntactic and semantic knowledge such as, for example, the SP-pattern `\texttt{agnt\_a NP \#NP \#agnt\_a act\_a V \#V \#act\_a pat\_a NP \#NP \#pat\_a obj\_a NP \#NP \#obj\_a}' in Figure \ref{ws_councilmen_fv_figure}, part of the discussion in Appendix \ref{syntax_and_semantics_appendix}.

\end{itemize}

In connection with the problem of finding the referent of a pronoun in a WSI, an SP-pattern like `texttt{PEACE\_LOVING PN \#PN v1 n0}' in column 21 in Figure \ref{ws_councilmen_fv_figure} is particularly important (Section \ref{disambiguation_of_they_section}), and likewise for the SP-pattern `texttt{PEACE\_LOVING PN \#PN FEARED VIOLENCE}' in column 21 in Figure \ref{ws_councilmen_fv_1_figure} (Section \ref{variations_section}).

The importance of SP-patterns like these is that they provide connections between things like `texttt{PEACE\_LOVING}', `texttt{PN \#PN}', `texttt{FEARED}', and `texttt{VIOLENCE}', which may be and usually are separated from each other by other things. It is precisely the ability to learn discontinuous dependencies which is required to learn those kinds of SP-patterns.

\subsubsection{Discussion}\label{discussion_section}

The method that is proposed in this paper for finding the referent for a pronoun in a WSI relies in part on a rather simple principle: that any person or group of people that is `peace loving' is also likely to `fear violence', and likewise for the association between a person or group having an interest in `radical change' and a willingness to `advocate revolution', and so on. The concept of SP-multiple-alignment within the SP Computer Model provides a means for this kind of association to exert its influence across intervening structures without disruption or disturbance by those intervening structures.

The essential simplicity of this idea seems to be in conflict with Terry Winograd's suggestion that resolving the ambiguity in each of his example sentences requires a ``sophisticated knowledge of councilmen, demonstrators, and politics'' \cite[p.~33]{winograd_1972}. Is this a genuine conflict or is there some way in which the two views may be reconciled? The suggestion here is that they can indeed be reconciled, something like this:

\begin{itemize}

    \item It is likely that, in real life, most adults will indeed have a sophisticated knowledge of councilmen, demonstrators, and politics, and  this knowledge will indeed be applied in the task of determining the referent of ``they'' in sentences like {\em The city councilmen refused the demonstrators a permit because they feared violence}. Their knowledge may, for example, contain many class-inclusion hierarchies like that shown in Figure \ref{ws_councilmen_fv_3_figure} in Section \ref{variations_section}.

    \item But in the same way that we use relatively short verbal labels for complex entities like `New York' or `Scotland', our non-verbal knowledge will contain many such labels in a `chunking-with-codes' method for achieving compression of information (\cite[Section 5]{sp_compression}, \cite[Section 2.2.8]{wolff_2006}).

    \item As we have seen in each of the examples presented in Section \ref{examples_section}, short identifiers are sufficient for determining the referent of the pronoun in each WSI. In more realistic examples, most such identifiers would each be a label or `code' for a relatively complex chunk of information.

    \item In short, the method that has been proposed in this paper for determining the referent of a pronoun in a WSI will work with short identifiers, but in more realistic settings, it is likely that most such identifiers would be associated with relatively complex bodies of knowledge.

\end{itemize}

\section{Conclusion}

In `Winograd Schema' kinds of sentence like {\em The city councilmen refused the demonstrators a permit because they feared violence} and {\em The city councilmen refused the demonstrators a permit because they advocated revolution} it is easy for people to decide what ``they'' refers to but it can be challenging for AI systems \cite[p.~33]{winograd_1972}. This paper describes with examples how, with the SP Theory of Intelligence and its realisation in the SP Computer Model, outlined in an Appendix, and with user-supplied information about relevant syntax and semantics, these kinds of problem of interpretation may be solved.

With the first sentence as an example, the key to finding what ``they'' refers to is: 1) to provide some information about one or more features of city councilmen such as, for example, that they value peace in their city; 2) to provide a means of recording the association between `peace loving' and `fear of violence'; and 3) to connect ``they'' into that association.

At present, this kind of information, and other relevant information about syntax and semantics, must be supplied to the SP System at the outset. But a comprehensive AI solution to these kinds of problem depends on an ability of the SP System, or any other kind of AI system, to learn relevant syntax and semantics for itself.

Existing strengths and weaknesses of the SP System in the unsupervised learning of English-like artificial languages suggest that it is not yet good enough to learn the kinds of syntactic and semantic knowledge that is needed to solve Winograd Schema types of problem, but there is clear potential for unsupervised learning in the SP System to be developed to a point where it can learn such knowledge for itself. In particular, there is clear potential for it to learn the kinds of association described under point (2) in the last-but-one paragraph---a kind of association which is similar to the kinds of discontinuous association in syntax---bridging intervening kinds of structure---which are found in most natural languages.

\section*{Appendices}

\begin{appendices}

\section{Outline of the SP system}\label{outline_of_sp_system_appendix}

The {\em SP System}---which means the {\em SP Theory of Intelligence} and its realisation in the {\em SP Computer Model}---is the product of an extended programme of research, seeking to simplify and integrate observations and concepts across artificial intelligence, mainstream computing, mathematics, and human learning, perception, and cognition, with information compression as a unifying theme. Despite its ambition, this objective has been largely met, as summarised in Appendix \ref{strengths_potential_of_sp_system_appendix}, below.

The SP System is described most fully in \cite{wolff_2006} and more briefly in \cite{sp_extended_overview}. Other information about the SP System may be found on \href{http://www.cognitionresearch.org/sp.htm}{www.cognitionresearch.org/sp.htm}.

\begin{figure}[!htbp]
\centering
\includegraphics[width=0.5\textwidth]{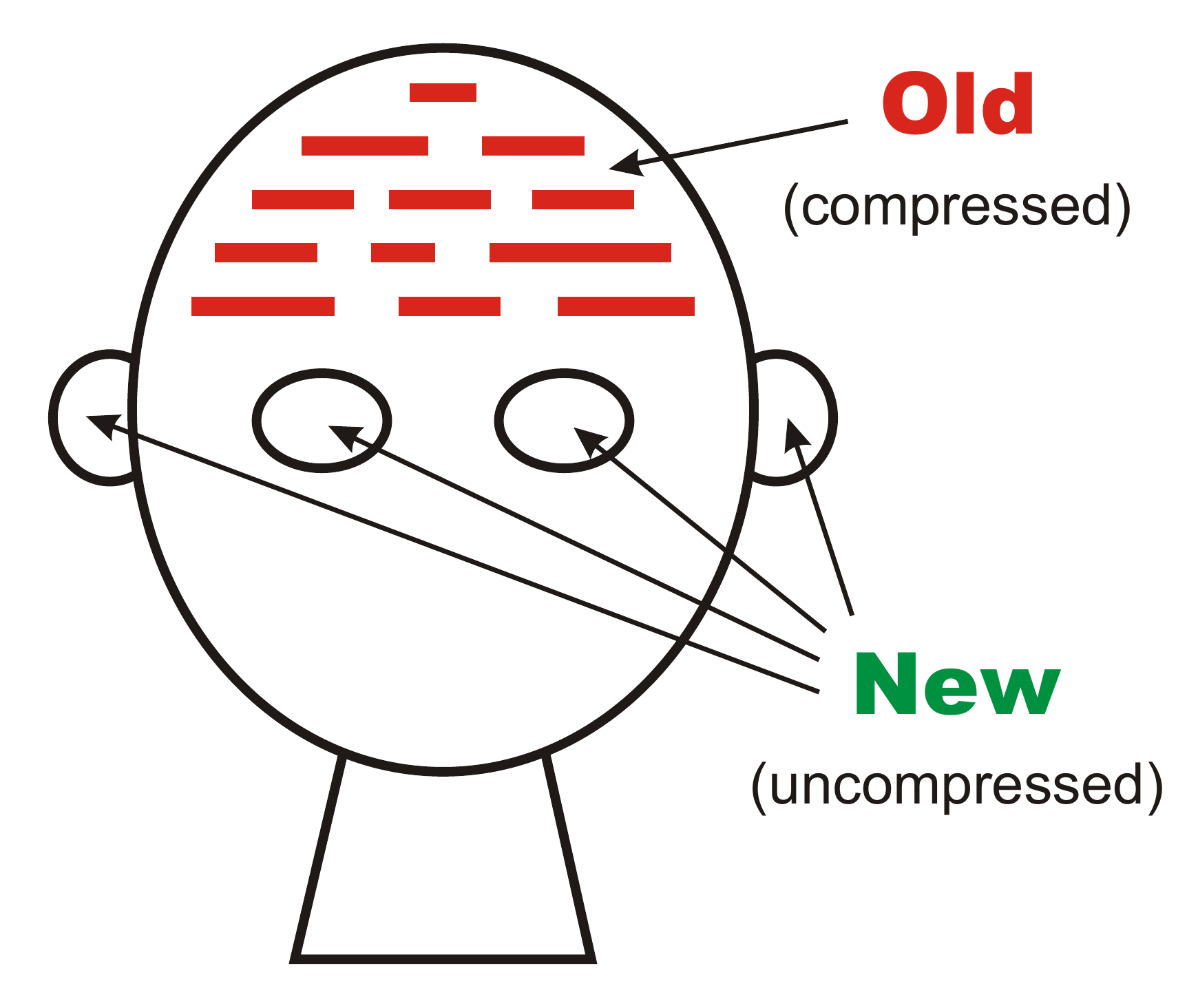}
\caption{Schematic representation of the SP System from an `input' perspective.}
\label{sp_input_perspective_figure}
\end{figure}

The SP System is conceived as a brain-like system that receives {\em New} information through its senses and stores some or all of it as {\em Old} information, as shown schematically in Figure \ref{sp_input_perspective_figure}.

In the SP System, all kinds of information or knowledge are represented in arrays of atomic {\em SP-symbols} in one or two dimensions called {\em SP-patterns}. At present, the SP Computer Model works only with one-dimensional SP-patterns but it is envisaged that it will be generalised to work with two-dimensional SP-patterns. And at present, an SP-symbol is an ASCII character, or a string of such characters bounded by spaces or the end of the SP-pattern in which it appears.

\subsection{Information compression, Ockham's razor, `simplicity', and `power'}\label{ic_simplicity_power_appendix}

Information compression, mentioned above, may be seen to be equivalent to a process of maximising the {\em Simplicity} of any given body of information, {\bf I}, by extraction of {\em redundancy} from {\bf I}, whilst retaining as much as possible of its descriptive {\em Power}. Hence the name `SP'.

Given that equivalence, information compression may be seen to have a two-fold significance in the SP programme of research:

\begin{itemize}

    \item Simplification and integration of observations and concepts across a broad canvass (the overarching goal of the SP research) is, in accordance with Ockham's razor, a process of developing a system that combines conceptual {\em Simplicity} of the system with high levels of explanatory or descriptive {\em Power}.

    \item Information compression is central in the workings of the SP System. In that connection, we can make a broad distinction between two kinds of processes in the SP System:

    \begin{itemize}

        \item {\em The building of SP-multiple-alignments}. The building of {\em SP-multiple-alignments} in the SP System, described in Section \ref{sp-multiple-alignment_appendix}, is a powerful means of compressing New information from the SP System's environment. It is also the key to the SP System's strengths in most aspects of intelligence, but with additional processing in unsupervised learning (next).

        \item {\em Unsupervised learning}. The process of unsupervised learning in the SP System (Appendix \ref{unsupervised_learning_appendix}) is another means of compressing information which incorporates the building of SP-multiple-alignments but which also includes a process of building one or more {\em SP-grammars}, meaning collections of Old SP-patterns which are relatively effective in the compression of a given set of New SP-patterns.

    \end{itemize}

\end{itemize}

The reason that information compression has been adopted as a unifying principle in the workings of the SP System is because of abundant evidence for the importance of information compression in human learning, perception, and cognition \cite{sp_compression}. The success of the system in modelling several aspects of human intelligence (Appendix \ref{versatility_in_intelligence_appendix}) provides corroborating evidence for the importance of information compression in human cognition.

A last point to mention in this connection is that it has been known for some time that there is an intimate relation between information compression and concepts of inference and probability \cite{shannon_weaver_1949,solomonoff_1964,solomonoff_1997,li_vitanyi_2014}. This makes it relatively straightforward for the SP System to calculate probabilities for inferences made by the system (\cite[Section 4.4]{sp_extended_overview}, \cite[Section 3.7]{wolff_2006}).

\subsection{SP-multiple-alignments}\label{sp-multiple-alignment_appendix}

The concept of {\em SP-multiple-alignment} in the SP System, which has been borrowed and adapted from the concept of `multiple sequence alignment' in bioinformatics, means the arrangement of two or more SP-patterns alongside each other, and the `stretching' of SP-patterns in a computer so that matching symbols are brought into line.

An example of an SP-multiple-alignment, created by the SP Computer Model, is shown in Figure \ref{fortune_brave_multiple_alignment_figure}. This is the best SP-multiple-alignment produced by the SP computer model with: a New SP-pattern, `\texttt{f o r t u n e f a v o u r s t h e b r a v e}' shown in column 0, representing a sentence to be parsed; and a repository of user-supplied Old SP-patterns representing grammatical categories, including morphemes and words. Some of those Old SP-patterns appear in columns 1 to 9 of the figure, one SP-pattern per column.

\newgeometry{top=2cm,left=2cm,right=2cm}

\begin{figure}[!htbp]
\fontsize{09.00pt}{10.80pt}
\centering
{\bf
\begin{BVerbatim}
0   1     2     3     4    5     6     7    8     9

                                 S
                                 0
                           NP -- NP
                           1
                           D
                           #D
                      N -- N
                      4
f ------------------- f
o ------------------- o
r ------------------- r
t ------------------- t
u ------------------- u
n ------------------- n
e ------------------- e
                      #N - #N
                           #NP - #NP
                VP ------------- VP
                3
          V --- V
          7
    Vr -- Vr
    6
f - f
a - a
v - v
o - o
u - u
r - r
    #Vr - #Vr
s ------- s
          #V -- #V
                NP ------------------------ NP
                                            1
                                       D -- D
                                       8
t ------------------------------------ t
h ------------------------------------ h
e ------------------------------------ e
                                       #D - #D
                                            N --- N
                                                  5
b ----------------------------------------------- b
r ----------------------------------------------- r
a ----------------------------------------------- a
v ----------------------------------------------- v
e ----------------------------------------------- e
                                            #N -- #N
                #NP ----------------------- #NP
                #VP ------------ #VP
                                 #S

0   1     2     3     4    5     6     7    8     9
\end{BVerbatim}
}
\caption{The best SP-multiple-alignment produced by the SP Computer Model with a New SP-pattern, `\texttt{f o r t u n e f a v o u r s t h e b r a v e}' (in column 0), representing a sentence to be parsed and a repository of user-supplied Old SP-patterns representing grammatical categories, including morphemes and words. Some of those Old SP-patterns appear in columns 1 to 9 of the SP-multiple-alignment, one SP-pattern per column.}
\label{fortune_brave_multiple_alignment_figure}
\end{figure}

\restoregeometry

Here, the meaning of `best' is that the given SP-multiple-alignment is the one which allows the New SP-pattern to be encoded most economically in terms of the Old SP-patterns in the SP-multiple-alignment, as described in \cite[Section 4.1]{sp_extended_overview} and \cite[Section 3.5]{wolff_2006}.

For any given New pattern and collection of Old SP-patterns, the number of possible SP-multiple-alignments is normally too large to be searched exhaustively. Hence, the SP Computer Model uses heuristic techniques which can normally find SP-multiple-alignments that are reasonably good, and may quite often be theoretically ideal, but which cannot guarantee to find such ideal SP-multiple-alignments.

\subsection{Unsupervised learning}\label{unsupervised_learning_appendix}

As mentioned in Appendix \ref{ic_simplicity_power_appendix} above, unsupervised learning in the SP System incorporates the building of SP-multiple-alignments but, in addition, there is a process of creating one or more {\em SP-grammars}, meaning collections of Old SP-patterns which are relatively good for the economical encoding of a given set of New SP-patterns.

The reason that, with respect to learning, the SP research is concentrating on the development of {\em unsupervised} learning---meaning learning without assistance from a `teacher' or anything equivalent---is because it appears that most human learning is unsupervised, and because of the belief that unsupervised learning can provide the foundation of other kinds of learning such as `supervised learning', `reinforcement learning', `learning by being told', `learning by imitation', and so on.

The main framework for unsupervised learning has been established in the SP Computer Model. It has already demonstrated an ability to discover generative grammars from unsegmented samples of English-like artificial languages, including segmental structures, classes of structure, and abstract patterns. But it has shortcomings summarised in \cite[Section 3.3]{sp_extended_overview}. How the system may be developed is outlined here:
\begin{itemize}

    \item {\em The learning of segmental structures}. The SP Computer Model has already demonstrated an ability to learn segmental structures such as words from data in which significant segments are not marked (Appendix \ref{unsupervised_learning_appendix}). Thus it has clear potential to learn such things as words and more abstract structures such as sentences. But a weakness of the model as it is now is that it does not learn structures at intermediate levels such as phrases and clauses ({\em ibid.}). It is anticipated that this problem will prove to be soluble.

    \item {\em The learning of class-inclusion structures}. Much the same may be said about the learning of class-inclusion structures such as nouns, verbs, adjectives, and so on. The SP Computer Model can learn such groupings at an abstract level but cannot learn structures at intermediate levels. As before, it is anticipated that this problem will prove to be soluble.

    \item {\em The learning of discontinuous dependencies}. Another shortcoming in how the SP Computer Model learns is that it cannot yet learn discontinuous dependencies in syntax such as gender dependencies (masculine or feminine) in a language like French (Appendix \ref{unsupervised_learning_appendix}). As before, it anticipated that this problem will prove soluble.

    \item {\em The learning of semantic structures}. An overarching goal in the SP programme of research is that structures and processes in the SP System should be applicable across a broad canvass including artificial intelligence, mainstream computing, mathematics, and human learning, perception, and cognition. In accordance with that principle, it is anticipated that, if the kinds of learning described in the previous three bullet points can be developed for the learning of syntax, they will prove to be applicable to the learning of 'semantic' structures, meaning knowledge of the non-syntactic world. The main change that will be needed will be generalisation of the SP System for SP-patterns in two dimensions, which will facilitate the learning of structures in three dimensions \cite[Sections 6.1 and 6.2]{sp_vision}. And there will need to progress in the learning of `commonsense' knowledge of the kinds described by \cite{davis_marcus_2015}.

    \item {\em The integration of syntax and semantics}. Probably the most challenging problem to be solved is to develop the SP System to a stage where it can learn plausible structures that integrate both natural language syntax and the non-syntactic knowledge that provides the semantics of natural language.

\end{itemize}

\subsection{SP-Neural}

{\em SP-Neural} is a version of the SP System expressed in terms of neurons and their inter-connections \cite{spneural_2016}. This is quite different from the popular `deep learning in artificial neural networks' \cite{schmidhuber_2015} and is likely to share the several advantages of the SP System compared with deep learning (\cite[Section V]{sp_alternatives}, \cite{spdlsol_2018}). SP-Neural has potential as a source of hypotheses in neuroscience and as a guide to explorations in that field.

\subsection{Strengths and potential of the SP System}\label{strengths_potential_of_sp_system_appendix}

In accordance with Ockham's razor (Appendix \ref{ic_simplicity_power_appendix}), the SP System combines conceptual {\em Simplicity} with high levels of descriptive or explanatory {\em Power}, the latter summarised in the subsections that follow. Further information may be found in \cite[Sections 5 to 12]{sp_extended_overview}, \cite[Chapters 5 to 9]{wolff_2006}, \cite{sp_alternatives}, and in other sources referenced in the subsections that follow.

\subsubsection{Versatility in aspects of intelligence}\label{versatility_in_intelligence_appendix}

In the modelling of human-like intelligence, strengths of the SP System includes: unsupervised learning, the analysis and production of natural language; pattern recognition that is robust in the face of errors in data; pattern recognition at multiple levels of abstraction; computer vision \cite{sp_vision}; best-match and semantic kinds of information retrieval; several kinds of reasoning (next paragraph); planning; and problem solving.

With regard to reasoning, the strengths of the SP system include: one-step `deductive' reasoning; chains of reasoning; abductive reasoning; reasoning with probabilistic networks and trees; reasoning with `rules'; nonmonotonic reasoning and reasoning with default values; Bayesian reasoning with `explaining away'; causal reasoning; reasoning that is not supported by evidence; the inheritance of attributes in class hierarchies and part-whole hierarchies. Where it is appropriate, probabilities for inferences may be calculated in a straightforward manner (\cite[Section 3.7]{wolff_2006}, \cite[Section 4.4]{sp_extended_overview}).

There is also potential in the system for spatial reasoning \cite[Section IV-F.1]{sp_autonomous_robots}, for what-if reasoning \cite[Section IV-F.2]{sp_autonomous_robots}, and for commonsense reasoning (this paper, and \cite{sp_csrk}).

\subsubsection{Versatility in the representation of knowledge}\label{versatility_in_representation_of_knowledge_appendix}

Although SP-patterns are not very expressive in themselves, they come to life in the SP-multiple-alignment framework. Within that framework, they may serve in the representation of several different kinds of knowledge, including: the syntax of natural languages; class-inclusion hierarchies (with or without cross classification); part-whole hierarchies; discrimination networks and trees; if-then rules; entity-relationship structures \cite[Sections 3 and 4]{wolff_sp_intelligent_database}; relational tuples ({\em ibid}., Section 3), and concepts in mathematics, logic, and computing, such as `function', `variable', `value', `set', and `type definition' (\cite[Chapter 10]{wolff_2006}, \cite[Section 6.6.1]{sp_benefits_apps}, \cite[Section 2]{sp_software_engineering}).

The addition of two-dimensional SP patterns to the SP computer model is likely to expand the representational repertoire of the SP system to structures in two-dimensions and three-dimensions, and the representation of procedural knowledge with parallel processing.

\subsubsection{Seamless integration of diverse aspects of intelligence, and diverse kinds of knowledge, in any combination}\label{seamless_integration_appendix}

An important third feature of the SP system, alongside its versatility in aspects of intelligence and its versatility in the representation of diverse kinds of knowledge, is that {\em there is clear potential for the SP system to provide seamless integration of diverse aspects of intelligence and diverse kinds of knowledge, in any combination.} This is because diverse aspects of intelligence and diverse kinds of knowledge all flow from a single coherent and relatively simple source: the SP-multiple-alignment framework.

It appears that seamless integration of diverse aspects of intelligence and diverse kinds of knowledge, in any combination, is {\em essential} in any artificial system that aspires to the fluidity, versatility and adaptability of the human mind.

\begin{figure}[!hbt]
\centering
\includegraphics[width=0.9\textwidth]{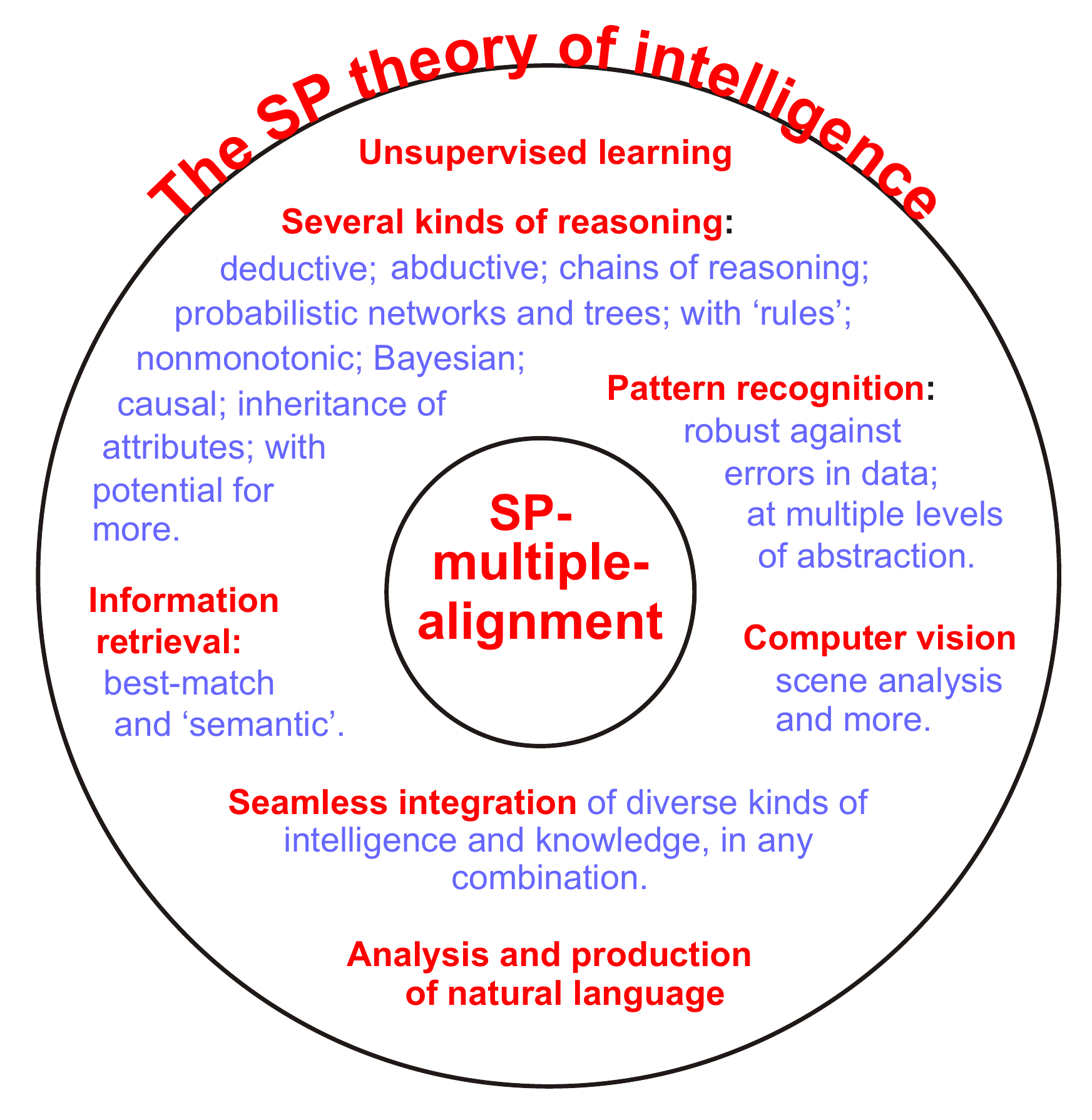}
\caption{A schematic representation of versatility and integration in the SP system, with SP-multiple-alignment centre stage.}
\label{versatility_integration_figure}
\end{figure}

Figure \ref{versatility_integration_figure} shows schematically how the SP system, with SP-multiple-alignment centre stage, exhibits versatility and integration.

\subsubsection{Potential benefits and applications of the SP system}\label{benefits_and_applications_appendix}

Apart from its strengths and potential in modelling aspects of human intelligence, it appears that the SP System has several potential benefits and applications \cite[Section 7]{sp_intro_2018}. These include: how the SP System may help solve nine problems with big data; the SP system opens up a radically new approach to the development of intelligence in autonomous robots; the SP System may form the basis for a database system with intelligence; the SP System may assist with medical diagnosis; the SP System may help in the development of computer vision and in the understanding of natural vision; SP-Neural may suggest hypotheses and avenues for research in neuroscience; the SP System has potential in the development of commonsense reasoning; other areas of application including \cite{sp_benefits_apps}: the simplification and integration of computing systems; applications of natural language processing; best-match and semantic forms of information retrieval; software engineering \cite{sp_software_engineering}; the representation of knowledge, reasoning, and the semantic web; information compression; bioinformatics; the detection of computer viruses; and data fusion; the concept of information compression via the matching and unification of patterns provides an entirely novel interpretation of mathematics \cite{sp_micmup} which is quite unlike anything described in existing writings about the philosophy of mathematics or its application in science and has several potential benefits and applications.

\section{Making connections between syntax and semantics}\label{syntax_and_semantics_appendix}

As mentioned in Section \ref{feared_violence_section}, the subject of this Appendix---the representation and processing of syntax and semantics in the example from that section and all other examples in this paper---is quite tentative and not relevant to the main substance of the paper---how WSIs may be disambiguated. Readers may safely ignore this Appendix if they wish.

By contrast with the SP-multiple-alignment in Figure \ref{fortune_brave_multiple_alignment_figure}, the SP-patterns in Figure \ref{ws_councilmen_fv_figure} attempt to show in a simplified way, how, with the SP System, syntactic structures may be associated with corresponding semantic structures. For example:

\begin{itemize}

    \item Each word is represented by an SP-pattern like this: `\texttt{N n3 councilmen COUNCILMEN PEACE\_LOVING \#N}' in column 14, where the lower-case string of characters `\texttt{councilmen}' is intended to represent the surface form of the word, while the associated string of characters in capital letters, `\texttt{COUNCILMEN}', is intended to represent, in a highly simplified form, the meaning of the word.

        The SP-symbol `\texttt{PEACE\_LOVING}', which is discussed in Section \ref{disambiguation_of_they_section}, may be seen to be a supplement to, or part of, the semantic information associated with `\texttt{councilmen}'.

        \sloppy The other symbols, `\texttt{N}', `\texttt{n3}', `\texttt{\#N}' in `\texttt{N n3 councilmen COUNCILMEN peace\_loving \#N}', are grammatical categories, much as in the SP-patterns in Figure \ref{fortune_brave_multiple_alignment_figure}.

    \item \sloppy The SP-pattern `\texttt{Sa agnt\_a NP \#NP \#agnt\_a act\_a V \#V \#act\_a pat\_a NP \#NP \#pat\_a obj\_a NP \#NP \#obj\_a actionA agentA patientA objectA \#Sa}' in column 7 is intended to represent both the syntax and semantics of `\texttt{the city councilmen refused the demonstrators a permit}'.

        Here, the first part of the pattern, `\texttt{agnt\_a NP \#NP \#agnt\_a act\_a V \#V \#act\_a pat\_a NP \#NP \#pat\_a obj\_a NP \#NP \#obj\_a}' is largely concerned with syntax, while `\texttt{actionA agentA patientA objectA}' is largely about semantics. Notice that the order of `\texttt{agnt\_a ...~\#agnt\_a act\_a ...~\#act\_a pat\_a ...~\#pat\_a obj\_a ...~\#obj\_a}' (meaning `agent, action, patient, object') in the `syntax' part is different from `\texttt{actionA agentA patientA objectA}' (`action, agent, patient, object') in the `semantics' part.

    \item In a similar way, the SP-pattern `\texttt{Sb agnt\_b NP \#NP \#agnt\_b act\_b V \#V \#act\_b obj\_b NP \#NP \#obj\_b actionB agentB objectB \#Sb}' in column 11 is intended to represent both the syntax and semantics of `\texttt{they feared violence}'. Here, the first part of the pattern, `\texttt{Sb agnt\_b NP \#NP \#agnt\_b act\_b V \#V \#act\_b obj\_b NP \#NP \#obj\_b}' (`agent', `action', `object') is largely concerned with syntax, while `\texttt{actionB agentB objectB}' (`action', `agent', `object') is largely about semantics, and the ordering is different in the two cases.

\end{itemize}

In English and many other languages, a given idea may often be expressed in more than one way. For example, the meaning of {\em John hit the ball} is essentially the same as {\em The ball was hit by John}. This suggests that there is a single semantic structure that may be expressed via two or more surface structures and that, for at least one of those surface structures, perhaps all of them, there is a process of mapping the surface structure on to the underlying semantic structure.

With the SP-multiple-alignment shown in Figure \ref{ws_councilmen_fv_figure}, that kind of mapping may be achieved like this:

\begin{itemize}

    \item With the SP-pattern in column 27, `\texttt{agnt\_a \#agnt\_a agentA}':

        \begin{itemize}

            \item The pair of SP-symbols `\texttt{agnt\_a \#agnt\_a}' in column 27 are aligned with matching symbols in column 7 that bridge the pair of symbols `\texttt{NP \#NP}' in that column, thus identifying the noun phrase `the city councilmen'.

            \item The SP-symbol `\texttt{agentA}' in column 27 is aligned with the matching symbol in column 7.

            \item The overall effect is to make a connection between `the city councilmen' and `\texttt{agentA}'.

        \end{itemize}

    \item There is a similar mapping of symbols: in `\texttt{act\_a \#act\_a actionA}' in column 28 to symbols in column 7; in `\texttt{pat\_a \#pat\_a patientA}' in column 23 to symbols in column 7; and in `\texttt{obj\_a \#obj\_a objectA}' in column 22 to symbols in column 7.

    \item The overall effect is to map the surface form `\texttt{the city councilmen refused the demonstrators a permit}' (`actor', `action', `patient', `object') into the form (`actionA', `agentA', `patientA', `objectA') in column 7, which may be regarded as a `semantic' version of the surface form.

    \item In a similar way, the surface form `\texttt{they feared violence}' may be seen to be mapped into the semantic form (`actionB', `agentB', `objectB') in column 11 via the SP-patterns `\texttt{agnt\_b \#agnt\_b agentB}' (column 25), `\texttt{act\_b \#act\_b actionB}' (column 26), and `\texttt{obj\_b \#obj\_b objectB}' (column 24).

\end{itemize}

\end{appendices}

\bibliographystyle{plain}

\begin{thebibliography}{10}

\bibitem{bailey_etal_2015}
D.~Bailey, A.~Harrison, Y.~Lierler, V.~Lifschitz, and J.~Michael.
\newblock The winograd schema challenge and reasoning about correlation.
\newblock In {\em Logical Formalizations of Commonsense Reasoning: Papers from
  the 2015 AAAI Spring Symposium}, pages 17--24, 2015.

\bibitem{davis_marcus_2015}
E.~Davis and G.~Marcus.
\newblock Commonsense reasoning and commonsense knowledge in artificial
  intelligence.
\newblock {\em Communications of the ACM}, 58(9):92--103, 2015.

\bibitem{emami_etal_2018}
A.~Emami, A.~Trischler, K.~Suleman, and J.~C.~K. Cheung.
\newblock A generalized knowledge hunting framework for the winograd schema
  challenge.
\newblock In {\em Proceedings of NAACL-HLT 2018: Student Research Workshop, New
  Orleans, Louisiana, June 2 - 4, 2018}, pages 25--31, 2018.

\bibitem{levesque_2011}
H.~J. Levesque.
\newblock The winograd schema challenge.
\newblock In {\em Proceedings of the Tenth International Symposium on Logical
  Formalizations of Commonsense Reasoning (Commonsense-2011)}, 2011.
\newblock Part of the AAAI Spring Symposium Series at Stanford University,
  March 21-23, 2011.

\bibitem{levesque_2014}
H.~J. Levesque.
\newblock On our best behaviour.
\newblock {\em Artificial Intelligence}, 212:27--35, 2014.

\bibitem{levesque_etal_2012}
H.~J. Levesque, E.~Davis, and L.~Morgenstern.
\newblock The winograd schema challenge.
\newblock In {\em Proceedings of the Thirteenth International Conference on
  Principles of Knowledge Representation and Reasoning}, pages 552--561, 2012.

\bibitem{li_vitanyi_2014}
M.~Li and P.~Vit{\'a}nyi.
\newblock {\em An Introduction to Kolmogorov Complexity and Its Applications}.
\newblock Springer, New York, 3rd edition, 2014.

\bibitem{morgenstern_ortiz_2015}
L.~Morgenstern and C.~L.~Ortiz Jr.
\newblock The winograd schema challenge: evaluating progress in commonsense
  reasoning.
\newblock In {\em Proceedings of the Twenty-Seventh Conference on Innovative
  Applications of Artificial Intelligence, 2015}, pages 4024--4025, 2015.

\bibitem{rahman_Ng_2012}
A.~Rahman and V.~Ng.
\newblock Resolving complex cases of definite pronouns: the {W}inograd {S}chema
  {C}hallenge.
\newblock In {\em Proceedings of the 2012 Joint Conference on Empirical Methods
  in Natural Language Processing and Computational Natural Language Learning},
  pages 777--789. Association for Computational Linguistics, 2012.

\bibitem{richard-bollans_etal_2018}
A.~Richard-Bollans, L.~G. {\'Alvarez}, and A.~G. Cohn.
\newblock The role of pragmatics in solving the winograd schema challenge.
\newblock In A.~S. Gordon, R.~Miller, and G.~Turan, editors, {\em Proceedings
  of the Thirteenth International Symposium on Commonsense Reasoning
  (Commonsense 2017), 06-08 Nov 2017, London, UK}, 2018.

\bibitem{schmidhuber_2015}
J.~Schmidhuber.
\newblock Deep learning in neural networks: an overview.
\newblock {\em Neural Networks}, 61:85--117, 2015.

\bibitem{schuller_2014}
P.~Sch{\"u}ller.
\newblock Tackling winograd schemas by formalizing relevance theory in
  knowledge graphs.
\newblock In {\em Proceedings of the Fourteenth International Conference on
  Principles of Knowledge Representation and Reasoning}, pages 358--367, 2014.

\bibitem{shannon_weaver_1949}
C.~E. Shannon and W.~Weaver.
\newblock {\em The Mathematical Theory of Communication}.
\newblock University of Illinois Press, Urbana, 1949.

\bibitem{sharma_etal_2015a}
A.~Sharma, N.~H. Vo, S.~Aditya, and C.~Baral.
\newblock Towards addressing the winograd schema challenge---building and using
  a semantic parser and a knowledge hunting module.
\newblock In {\em Proceedings of the Twenty-Fourth International Joint
  Conference on Artificial Intelligence (IJCAI 2015)}, pages 1319--1325, 2015.

\bibitem{sharma_etal_2015b}
A.~Sharma, N.~H. Vo, S.~Gaur, and C.~Baral.
\newblock An approach to solve winograd schema challenge using automatically
  extracted commonsense knowledge.
\newblock In {\em Logical Formalizations of Commonsense Reasoning: Papers from
  the 2015 AAAI Spring Symposium}, pages 141--144, 2015.

\bibitem{solomonoff_1964}
R.~J. Solomonoff.
\newblock A formal theory of inductive inference. {P}arts {I} and {II}.
\newblock {\em Information and Control}, 7:1--22 and 224--254, 1964.

\bibitem{solomonoff_1997}
R.~J. Solomonoff.
\newblock The discovery of algorithmic probability.
\newblock {\em Journal of Computer and System Sciences}, 55(1):73--88, 1997.

\bibitem{winograd_1972}
T.~Winograd.
\newblock Understanding natural language.
\newblock {\em Cognitive Psychology}, 3(1):1--191, 1972.

\bibitem{wolff_2006}
J.~G. Wolff.
\newblock {\em Unifying Computing and Cognition: the {SP} Theory and Its
  Applications}.
\newblock CognitionResearch.org, Menai Bridge, 2006.
\newblock {ISBN}s: 0-9550726-0-3 (ebook edition), 0-9550726-1-1 (print
  edition). Distributors, including Amazon.com, are detailed on
  \href{http://bit.ly/WmB1rs}{bit.ly/WmB1rs}.

\bibitem{wolff_sp_intelligent_database}
J.~G. Wolff.
\newblock Towards an intelligent database system founded on the {SP} theory of
  computing and cognition.
\newblock {\em Data \& Knowledge Engineering}, 60:596--624, 2007.
\newblock arXiv:cs/0311031 [cs.DB],
  \href{http://bit.ly/1CUldR6}{bit.ly/1CUldR6}.

\bibitem{sp_extended_overview}
J.~G. Wolff.
\newblock The {SP} theory of intelligence: an overview.
\newblock {\em Information}, 4(3):283--341, 2013.
\newblock arXiv:1306.3888 [cs.AI],
  \href{http://bit.ly/1NOMJ6l}{bit.ly/1NOMJ6l}.

\bibitem{sp_vision}
J.~G. Wolff.
\newblock Application of the {SP} theory of intelligence to the understanding
  of natural vision and the development of computer vision.
\newblock {\em SpringerPlus}, 3(1):552--570, 2014.
\newblock arXiv:1303.2071 [cs.CV],
  \href{http://bit.ly/2oIpZB6}{bit.ly/2oIpZB6}.

\bibitem{sp_autonomous_robots}
J.~G. Wolff.
\newblock Autonomous robots and the {SP} theory of intelligence.
\newblock {\em IEEE Access}, 2:1629--1651, 2014.
\newblock arXiv:1409.8027 [cs.AI],
  \href{http://bit.ly/18DxU5K}{bit.ly/18DxU5K}.

\bibitem{sp_benefits_apps}
J.~G. Wolff.
\newblock The {SP} theory of intelligence: benefits and applications.
\newblock {\em Information}, 5(1):1--27, 2014.
\newblock arXiv:1307.0845 [cs.AI],
  \href{http://bit.ly/1FRYwew}{bit.ly/1FRYwew}.

\bibitem{sp_csrk}
J.~G. Wolff.
\newblock Commonsense reasoning, commonsense knowledge, and the {SP} theory of
  intelligence.
\newblock Technical report, CognitionResearch.org, 2016.
\newblock Submitted for publication, arXiv:1609.07772 [cs.AI],
  \href{http://bit.ly/2eBoE9E}{bit.ly/2eBoE9E}.

\bibitem{spneural_2016}
J.~G. Wolff.
\newblock Information compression, multiple alignment, and the representation
  and processing of knowledge in the brain.
\newblock {\em Frontiers in Psychology}, 7:1584, 2016.
\newblock arXiv:1604.05535 [cs.AI],
  \href{http://bit.ly/2esmYyt}{bit.ly/2esmYyt}.

\bibitem{sp_alternatives}
J.~G. Wolff.
\newblock The {SP} theory of intelligence: its distinctive features and
  advantages.
\newblock {\em IEEE Access}, 4:216--246, 2016.
\newblock arXiv:1508.04087 [cs.AI],
  \href{http://bit.ly/2qgq5QF}{bit.ly/2qgq5QF}.

\bibitem{sp_compression}
J.~G. Wolff.
\newblock Information compression via the matching and unification of patterns
  as a unifying principle in human learning, perception, and cognition.
\newblock Technical report, CognitionResearch.org, 2017.
\newblock Submitted for publication.
  \href{http://bit.ly/2ruLnrV}{bit.ly/2ruLnrV}, viXra:1707.0161v2,
  hal-01624595, v1.

\bibitem{sp_software_engineering}
J.~G. Wolff.
\newblock Software engineering and the {SP} theory of intelligence.
\newblock Technical report, CognitionResearch.org, 2017.
\newblock Submitted for publication, arXiv:1708.06665 [cs.SE],
  \href{http://bit.ly/2w99Wzq}{bit.ly/2w99Wzq}.

\bibitem{sp_intro_2018}
J.~G. Wolff.
\newblock Introduction to the {SP} theory of intelligence.
\newblock Technical report, CognitionResearch.org, 2018.
\newblock arXiv:1802.09924, \href{http://bit.ly/2ELq0Jq}{bit.ly/2ELq0Jq}.

\bibitem{sp_micmup}
J.~G. Wolff.
\newblock Mathematics as information compression via the matching and
  unification of patterns.
\newblock Technical report, CognitionResearch.org, 2018.
\newblock Submitted for publication. arXiv:1808.07004 [cs.AI],
  \href{https://bit.ly/2LWbjtK}{bit.ly/2LWbjtK}.

\bibitem{spdlsol_2018}
J.~G. Wolff.
\newblock Solutions to problems with deep learning.
\newblock Technical report, CognitionResearch.org, 2018.
\newblock arXiv:1801.05457 [cs.LG],
  \href{http://bit.ly/2AJzu4j}{bit.ly/2AJzu4j}.

\end{thebibliography}

\end{document}